\pdfoutput=1
\pdfoutput=1
\pdfoutput=1
\pdfoutput=1
\pdfoutput=1
\documentclass[acmtog,screen,author]{acmart}
\AtBeginDocument{%
  \providecommand\BibTeX{{%
    \normalfont B\kern-0.5em{\scshape i\kern-0.25em b}\kern-0.8em\TeX}}}

\setcopyright{acmcopyright}
\copyrightyear{2023}
\acmYear{2023}
\acmDOI{XXXXXXX.XXXXXXX}

\acmSubmissionID{439}

\usepackage{stfloats}
\usepackage{float} 
\usepackage{subfigure}  
\usepackage{algorithm}
\usepackage{algorithmic}
\usepackage{bm}
\usepackage{multirow}
\newcommand{\mbomega} {\boldsymbol{\omega}}
\newcommand{\mbmu} {\boldsymbol{\mu}}

\newcommand{\ff}[1]{{#1}}

\begin{document}

\setcopyright{acmlicensed}
\acmJournal{TOG}
\acmYear{2023} \acmVolume{42} \acmNumber{6} \acmArticle{1} \acmMonth{12} \acmPrice{15.00}\acmDOI{10.1145/3618360}

\title{Manifold Path Guiding for Importance Sampling Specular Chains}


\author{Zhimin Fan}
\email{zhiminfan2002@gmail.com}
\affiliation{%
  \institution{State Key Lab for Novel Software Technology, Nanjing University}
  \city{Nanjing}
  \country{China}
}
\authornote{Joint first authors.}
\author{Pengpei Hong}
\email{hpommpy@gmail.com}
\affiliation{%
  \institution{University of Utah}
  \city{Salt Lake City} % (?)
  \country{United States of America}
}
\authornotemark[1]
\authornote{The work was done at Nanjing University.}
\author{Jie Guo}
\email{guojie@nju.edu.cn}
\affiliation{%
  \institution{State Key Lab for Novel Software Technology, Nanjing University}
  \city{Nanjing}
  \country{China}
}
\authornote{Corresponding authors.}
\author{Changqing Zou}
\email{changqing.zou@zju.edu.cn}
\affiliation{%
  \institution{Zhejiang Lab and State Key Lab of CAD\&CG, Zhejiang University}
  \city{Hangzhou}
  \country{China}
}
\author{Yanwen Guo}
\email{ywguo@nju.edu.cn}
\affiliation{%
  \institution{State Key Lab for Novel Software Technology, Nanjing University}
  \city{Nanjing}
  \country{China}
}
\authornotemark[3]
\author{Ling-Qi Yan}
\email{lingqi@cs.ucsb.edu}
\affiliation{%
  \institution{University of California, Santa Barbara}
  \city{Santa Barbara}
  \country{United States of America}
}

% \renewcommand{\shortauthors}{Zhimin Fan, Pengpei Hong, Jie Guo, Changqing Zou, Yanwen Guo, and Ling-Qi Yan}

\begin{teaserfigure}
  \includegraphics[width=1\textwidth]{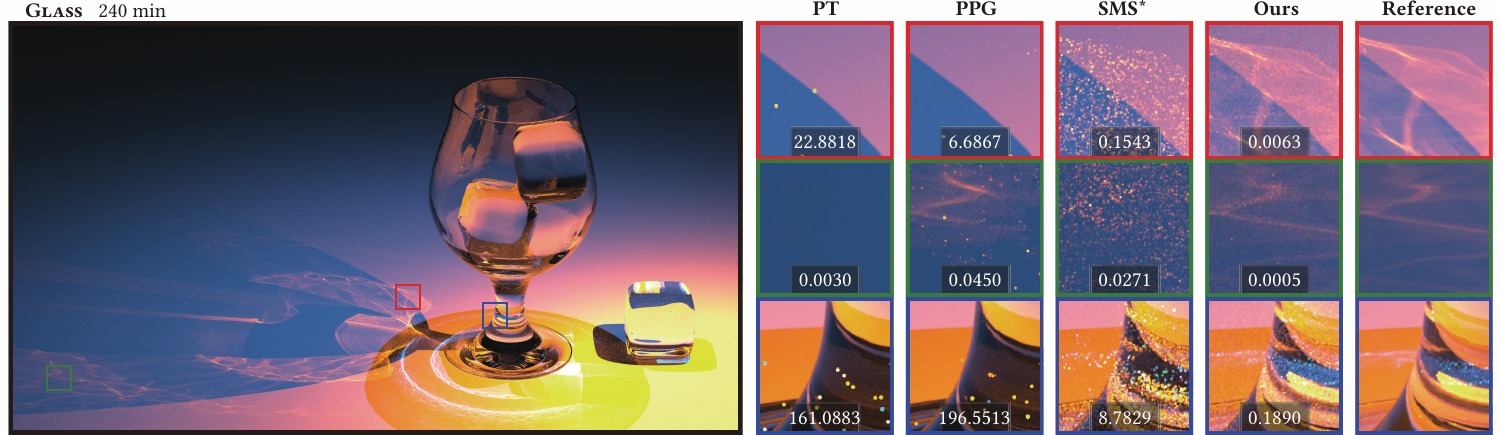}  % WIP
  \caption{Rendering a complex scene involving challenging light paths with multiple consecutive specular bounces. The long specular chains ($>5$ in this scene) create great obstacles to existing path sampling algorithms, while the proposed \emph{manifold path guiding} method addresses this issue, faithfully reproducing high-frequency caustics and noticeably reducing the variance. Here, we compare our method with Path Tracing (PT), Practical Path Guiding (PPG) \cite{Muller17, mueller19guiding}, and an extension (supporting various chain types) of Specular Manifold Sampling (SMS*) \cite{Zeltner20} at the same rendering time. 
  Quantitative error in terms of MSE is reported for each closeup.}
  \Description{...}
  \label{fig:teaser}
\end{teaserfigure}

\begin{abstract}
Complex visual effects such as caustics are often produced by light paths containing multiple consecutive specular vertices (dubbed \emph{specular chains}), which pose a challenge to unbiased estimation in Monte Carlo rendering.
In this work, we study the light transport behavior within a sub-path that is comprised of a specular chain and two non-specular separators. 
We show that the specular manifolds formed by all the sub-paths could be exploited to provide coherence among sub-paths. By reconstructing continuous energy distributions from historical and coherent sub-paths, seed chains can be generated in the context of importance sampling and converge to admissible chains through manifold walks. We verify that importance sampling the seed chain in the continuous space reaches the goal of importance sampling the discrete admissible specular chain. 
Based on these observations and theoretical analyses, a progressive pipeline, \emph{manifold path guiding}, is designed and implemented to importance sample challenging paths featuring long specular chains. To our best knowledge, this is the first general framework for importance sampling discrete specular chains in regular Monte Carlo rendering. Extensive experiments demonstrate that our method outperforms state-of-the-art unbiased solutions with up to 40× variance reduction, especially in typical scenes containing long specular chains and complex visibility.

\end{abstract}




\maketitle

\section{Introduction}
Monte Carlo (MC) integration using stochastic samples has long been the de facto solution to the problem of physically-based light transport simulation \cite{Christensen16, Fascione18, Keller15}. Over the past decades, great efforts have been devoted to significantly improving the convergence rate and reducing the noise for MC-based rendering algorithms.

However, several common scenes still lack effective and robust sampling strategies to handle certain classes of light paths well. In particular, long paths with multiple consecutive specular or near-specular scattering events (widely appearing in scenes with curved metals, water, and glasses as shown in Fig. \ref{fig:teaser}) have always been a nightmare for existing physically-based rendering engines, causing significantly slow convergence. The key difficulty lies in finding valid path samples that are not only ``important'' but also satisfy all physical constraints at specular reflective/refractive vertices \cite{Jakob12}.

To reduce the variance for scenes with multiple specular scattering events, many approaches have been tried. 
Early attempts leverage Metropolis Light Transport (MLT) \cite{Veach97MLT} to explore high-energy regions in the path space. However, even if some specific mutation strategies have been designed to handle paths containing specular chains \cite{Jakob12, Kaplanyan2014}, it is still difficult to find the so-called Specular-Diffuse-Specular (SDS) paths.
Another line of work leverage fitted energy distributions to directly sample the difficult paths \cite{Vorba14, Muller17}. Despite their success in production, they fail to handle pure specular cases (i.e., specular paths adjacent to point light sources). 

Recently, \citet{Zeltner20} proposed Specular Manifold Sampling (SMS), a general sampling strategy tailored for specular light paths. Unfortunately, this method favors short specular paths with one or two vertices, and its performance reduces dramatically when the paths become longer. Moreover, SMS samples specular paths according to the size of convergence basins and ignores their energy distributions which are critical for variance reduction. 
Until now, importance sampling arbitrarily long \emph{specular chains} (i.e., light paths containing multiple consecutive specular vertices) is still a far less explored field in computer graphics.

Our goal in this paper is to analyze the best possible importance sampling strategy for specular chains and find a practical solution to approach it without incurring significant overhead to the existing rendering pipelines. 
Due to the discrete nature of the \emph{admissible specular chain} space and its high dimensionality, conducting importance sampling directly in that space may be difficult.
To address this key issue, we exploit a continuous space in which seed chains can be importance sampled and converge to corresponding admissible chains via manifold walk \cite{Jakob12}.
We further reduce the dimensionality of the space by in-depth analyzing the light transport behavior of admissible specular chains, thus making the task tractable and easily supporting glossy cases.

We design and implement a practical rendering pipeline, named \emph{manifold path guiding}, to construct continuous distributions for importance sampling seed chains. After initialization, the pipeline progressively learns specular manifolds via reconstructing continuous spatial-directional distributions from historical sub-paths, generating new seed chains following the distributions, and performing manifold walks to obtain new admissible sub-paths. The learned distributions converge after several iteration steps and are then used in the final rendering stage. We show through extensive experiments that the proposed method allows us to faithfully produce complex caustics stemming from arbitrarily long specular chains, and it
outperforms previous unbiased MC solutions (including those tailored for specular paths) with more than one order of magnitude variance reduction in equal rendering time.

In summary, our contributions are:
\begin{enumerate}
    \item A formulation of the task of specular chains importance sampling that covers any number of consecutive specular vertices,
    \item A general solution that obtains and utilizes a continuous distribution exploiting the continuity of specular manifold,
    \item \ff{A practical pipeline} that progressively gathers sub-path samples, achieving fast convergence even for paths with long specular chains and complex visibility.
\end{enumerate}

\section{Related Work}

\paragraph{Bidirectional Monte Carlo and MCMC methods} 
Importance sampling difficult paths has been a long-standing challenge in rendering. Bidirectional approaches \cite{Lafortune93BDPT, Veach95BDPT} try to solve this problem by sampling from the viewpoint and the emitters independently and then properly connecting sub-paths. However, they fail to handle the SDS paths.

Biased methods, such as photon mapping \cite{Jensen95PM, Hachisuka09SPPM} and regularization \cite{Kaplanyan2013PathSR, Jendersie2019MicrofacetMR, Weier2021OPSR}, are also designed to find difficult paths for caustics. Due to their biased nature, these methods tend to lose details from high-frequency caustics.

Another family of Monte Carlo methods samples difficult paths by running a Markov chain of paths, known as MCMC approaches \cite{Sik2020SurveyOM}.
Pioneered by Metropolis Light Transport (MLT) \cite{Veach97MLT}, many efforts have been devoted to significantly enhancing the performance of path mutations \cite{Jakob12}. 
Although MCMC-based methods reduce variance significantly, they are plagued with splotchy non-uniform noise and temporal flickering artifacts in animations. 

Unlike these methods, our approach works in a regular and unbiased Monte Carlo manner. 
We stride over the limitation by connecting chains between separators. This belongs to the non-local sampling strategies \cite{Veach98}.

\paragraph{Path guiding.} Monte Carlo rendering techniques rely heavily on importance sampling when constructing light transport paths. So far, the most promising sampling distributions are obtained based on learned scene priors \cite{Vorba19}. 

Existing works have studied various distribution representations. For instance, \citet{Lafortune95} stored incident radiance of previously traced rays in a 5D tree, and \citet{Jensen95G} estimated histograms of incoming radiance from photons. Gaussian Mixture Models (GMM) \cite{Vorba14} are found to be flexible, while quad-trees \cite{Muller17} succeed in practice due to their adaptability. 

Some approaches further take consideration of the correlation between consecutive vertices in the full path, e.g., using product importance sampling \cite{Herholz2016ProductIS}, parallax-aware warping \cite{Ruppert2020RobustFO} and spatial correlation \cite{Dodik21, Schssler2022PathGW}.
Recently, \citet{Li22} utilized representative specular paths to enable effectively guided rendering of caustics, but their methods struggled to support specular or low-roughness materials. 

Guided path sampling achieves great success when high-frequency details are absent in the radiance distribution. However, when faced with paths containing multiple specular interactions, the radiance distribution usually involves high-frequency variations that bottleneck existing guiding techniques \cite{Loubet20}. 
We address this issue by importance sampling the specular chains directly. 
The proposed method allows us to find admissible paths in a very efficient way and enables the creation of high-frequency caustics from pure specular surfaces with low variance.

\paragraph{Specular light transport.} Due to the intrinsic difficulties in specular light transport, many specialized methods have been proposed to sample challenging specular paths. 

A line of approaches performs exhaustive searching and root-solving to find all specular chains connecting two endpoints \cite{Mitchell92}. Due to the high computational complexity, these methods either only work for one specular bounce \cite{Walter09, Loubet20} or incur significant performance degradation as the number of bounces increases \cite{Wang20}.

Other methods lower the computational burden by stochastical sampling at the cost of introducing variance.
Manifold Exploration Metropolis Light Transport (MEMLT) \cite{Jakob12} allows random walks on a specular manifold with Newton's method.
This, as a pioneer work in this field, is extended to half-vector space \cite{Kaplanyan2014} and then introduced to regular Monte Carlo sampling as Manifold Next-Event Estimation (MNEE) \cite{Hanika15}. 
MNEE heuristically generates a deterministic initial chain connecting the shading point to the light source, followed by a manifold walk for a feasible solution. 
With fixed initialization, MNEE can only find at most one specular chain connecting a given pair of endpoints, resulting in energy loss.
\begin{figure}[tbp]
  \centering
  \includegraphics[width=1.0\linewidth]{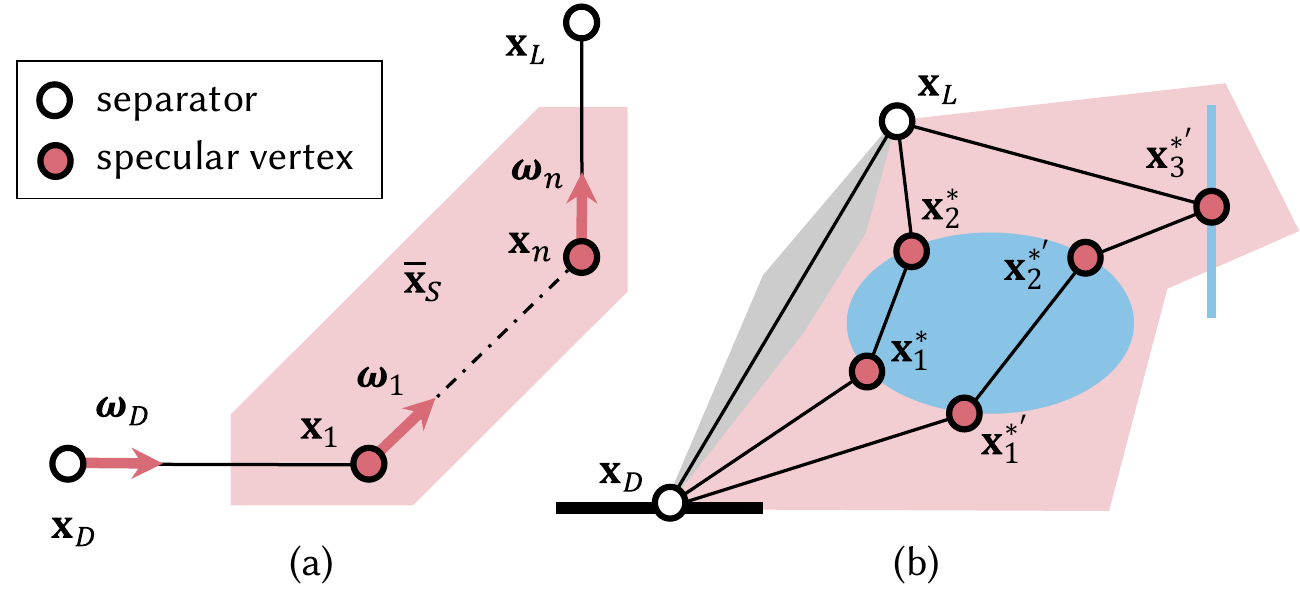}
  \caption{\textbf{(a)} \textbf{Notations for a specular chain} $\overline{\mathbf{x}}_S$ (in the red-shaded area) which is comprised of $n$ specular vertices $\mathbf{x}_1,\mathbf{x}_2,…,\mathbf{x}_n$. By connecting two separators $\mathbf{x}_D$ and $\mathbf{x}_L$ with a specular chain $\overline{\mathbf{x}}_S$, we produce a sub-path $(\mathbf{x}_D, \overline{\mathbf{x}}_S, \mathbf{x}_L)$. \textbf{(b) Throughput between two separators}, which comes from both direct and indirect connections. In this example, the indirect part is comprised of two admissible chains $\overline{\mathbf{x}}_S^* = \mathbf{x}_1^*, \mathbf{x}_2^*$ and $\overline{\mathbf{x}}_S^{*}{}' = {\mathbf{x}_1^*}', {\mathbf{x}_2^*}', {\mathbf{x}_3^*}'$. \ff{Thus, the admissible chain space $\mathcal{P}_S^* = \{\overline{\mathbf{x}}_S^*, \overline{\mathbf{x}}_S^*{}'\}$. }
  } 
  \Description{...}
  \label{fig_fullpath}
\end{figure}
Specular Manifold Sampling (SMS) \cite{Zeltner20} addressed the above issue with random initialization. Accompanied by an unbiased reciprocal probability estimator, it is able to preserve energy for complex caustics. Unfortunately, the simple treatment of initial guesses by uniform sampling will cause a high variance.

In comparison, we take the energy contribution of each chain into consideration during sampling, forming a new and generic importance sampling paradigm suitable for a wide range of specular chains, including very long chains and chains with complex visibility.

\section {Background}

We first review some background knowledge regarding our topic.

\paragraph {Paths with specular chains}

\citet{Veach98} introduced \textit{non-local sampling} that evaluates the full path integral in Monte Carlo rendering by sampling separators (i.e., diffuse\footnote{We utilize Heckbert's notation \cite{Heckbert90} for path classification: a full path is described by regular expressions in the form $E(S|D)^*L$, with each letter representing a vertex of a path. 
For ease of discussion, we assume that each path is comprised of pure specular and diffuse vertices. The extension to glossy will be discussed later.} vertices) first and then connecting specular chains (i.e., chains of consecutive specular vertices) between adjacent separators. 
Hence, importance sampling a full path boils down to importance sampling separators and chains.

While there are existing approaches (e.g., irradiance caching \cite{Ward1988ART}) for separator sampling, connecting chains is still a challenge and is our major concern in this paper.

Without loss of generality, we denote $\mathbf{x}_D$ and $\mathbf{x}_L$ as two separators, where $\mathbf{x}_D$ is a diffuse vertex and $\mathbf{x}_L$ is on the light source. $\overline {\mathbf{x}}_S$ represents a chain connecting $\mathbf{x}_D$ and $\mathbf{x}_L$, which is comprised of specular vertices $\mathbf{x}_1, \mathbf{x}_2, ..., \mathbf{x}_n$. 
We call $(\mathbf{x}_D, \mathbf{x}_L)$ as a \textit{configuration} and $(\mathbf{x}_D, \overline{\mathbf{x}}_S, \mathbf{x}_L)$ as a \textit{sub-path}.
For brevity, we also refer to $\mathbf{x}_D$ as $\mathbf{x}_0$, $\mathbf{x}_L$ as $\mathbf{x}_{n+1}$. These symbols are demonstrated in Fig. \ref{fig_fullpath}(a) and summarized in Table \ref{table1}.

\paragraph {Specular manifolds}
To satisfy all physically-based specular constraints along a path, existing solutions mostly build on top of \citet{Jakob12}'s path space manifold.
In this context, an \textit{admissible chain} $\overline{\mathbf{x}}_S^*$ \cite{Hanika15} refers to a specular chain that satisfies all the constraints, which could have a non-zero throughput once it passes the visibility test. 
In general, we could safely assume that there is a discrete and finite set of admissible chains connecting two separators. Although infinite cases theoretically exist, they rarely occur in natural rendering \cite{Wang20, Zeltner20} as discussed in our supplemental document.

\begin{table}
  \caption{List of important notations.}
  \label{table1}
  \begin{tabular}{cl}
    \toprule
    \textbf{Symbol} & \textbf{Definition} \\
    \midrule
    $\overline{\mathbf{x}}_S, \overline{\mathbf{x}}_S^*$ & Specular chain, admissible chain\\
    $(\mathbf{x}_D,\overline{\mathbf{x}}_S, \mathbf{x}_L)$ & Sub-path \\
    $\mathbf{x}_1,...,\mathbf{x}_n$ & Specular vertices of $\overline{\mathbf{x}}_S$ \\
    $\mbomega_D, \mbomega_D^*$ & Direction from $\mathbf{x}_D$ to $\mathbf{x}_1$, $\mathbf{x}_1^*$  \\
    $\tau_1, ..., \tau_n$ & Type of the scattering event of each vertex \\
    $T(\mathbf{x}_D, \overline{\mathbf{x}}_S^*, \mathbf{x}_L)$ & Sub-path throughput \\
  \bottomrule
\end{tabular}
\end{table}

\paragraph{Throughput between separators}
Between two separators $\mathbf{x}_D$ and $\mathbf{x}_L$, we use the term \textit{throughput} to refer to the irradiance received at $\mathbf{x}_D$ contributed by the unit-area surface at $\mathbf{x}_L$. 
The throughput can be subdivided into two independent parts \cite{Veach98}. The first part comes from the direct connection between $\mathbf{x}_D$ and $\mathbf{x}_L$, which can be trivially evaluated and thus will not be considered in the following discussions. 
The second part corresponds to all connections passing through at least one intermediate specular vertex. We formulate it as
$\sum_{\overline{\mathbf{x}}_S^* \in \mathcal{P}^*_S} T (\mathbf{x}_D,\overline{\mathbf{x}}^*_S, \mathbf{x}_L)
\label{Eq:TiSum}
$, a summation of all admissible chains,
\ff{where the admissible chain space $\mathcal{P}^*_S=\mathcal{P}^*_S (\mathbf{x}_D, \mathbf{x}_L)$ is a finite, countable set containing all admissible chains connecting $\mathbf{x}_D$ and $\mathbf{x}_L$ as shown in Fig. \ref{fig_fullpath}(b).}
The \textit{sub-path throughput} $T({\mathbf{x}}_D,\overline{{\mathbf{x}}}_S^*, {\mathbf{x}}_L)$ is part of the throughput contributed by a specified admissible chain $\overline{{\mathbf{x}}}_S^*$. Its value is determined by the product of several terms \cite{Jakob2013LightTO, Hanika15}:
\begin{equation}
T (\mathbf{x}_D,\overline{\mathbf{x}}_S^*, \mathbf{x}_L)=
\mathbf{\kappa}(\mathbf{x}_D,\overline{\mathbf{x}}_S^*, \mathbf{x}_L)
G(\mathbf{x}_D, \overline{\mathbf{x}}_S^*, \mathbf{x}_L)
{L_o(\mathbf{x}_n^*, \mathbf{x}_L)}
\label{Eq:ThroughputEval}
\end{equation}
with $\kappa$ being a unitless specular scattering value that folds the reflection/refraction coefficients at each specular vertex (e.g., the Fresnel term for reflection), $G$ denoting the Generalized Geometric Term (GGT) \cite{Jakob12, Wang20} which relates the differential solid angle at $\mathbf{x}_D$ to the differential surface area at $\mathbf{x}_L$ (including the visibility function) and $L_o(\mathbf{x}_n^*, \mathbf{x}_L)$ representing the outgoing radiance at $\mathbf{x}_L$ towards $\mathbf{x}_n^*$.

\paragraph{Manifold walks and seed chain sampling}

Existing solutions of specular path sampling \cite{Kaplanyan2014, Hanika15, Zeltner20} are mostly built on the top of \emph{manifold walks} \cite{Jakob12}. Through combining Newton's method and a reprojection step, it forms a general and efficient solver that can reach an admissible chain from a \textit{seed chain} $\overline{\mathbf{x}}_S$ (i.e., the initial state of the solver). 

\ff{In general, through the solver, discrete admissible chains in the discrete space $\mathcal{P}_S^*$ can be indirectly sampled by sampling a seed chain first from a continuous probability distribution $p(\overline{\mathbf{x}}_S | \mathbf{x}_D, \mathbf{x}_L)$\footnote{\ff{We use an upper-case $P$ to denote discrete probability mass functions and use a lower-case $p$ to represent continuous probability density functions.}} in the continuous specular chain space $\mathcal{P}_S$ (space formed by all chains of specular vertices) and then solve for an admissible chain via manifold walks. }A conditional probability distribution $P(\overline{\mathbf{x}}_S^* | \overline{\mathbf{x}}_S, \mathbf{x}_D, \mathbf{x}_L)$ demonstrates the convergence behavior of manifold walk. 
\ff{This also defines the \emph{convergence basin} \cite{Zeltner20}:
\begin{equation}
\mathcal{B}(\overline{\mathbf{x}}_S^* | \mathbf{x}_D, \mathbf{x}_L) = 
\{
\overline{\mathbf{x}}_S \mid
P(\overline{\mathbf{x}}_S^* | \overline{\mathbf{x}}_S, \mathbf{x}_D, \mathbf{x}_L) = 1
\}
\mathrm{.}
\end{equation}}

\begin{figure}[tbp]
\centering
	\begin{minipage}{1.00\linewidth}
  			\includegraphics[width=1.0\linewidth]{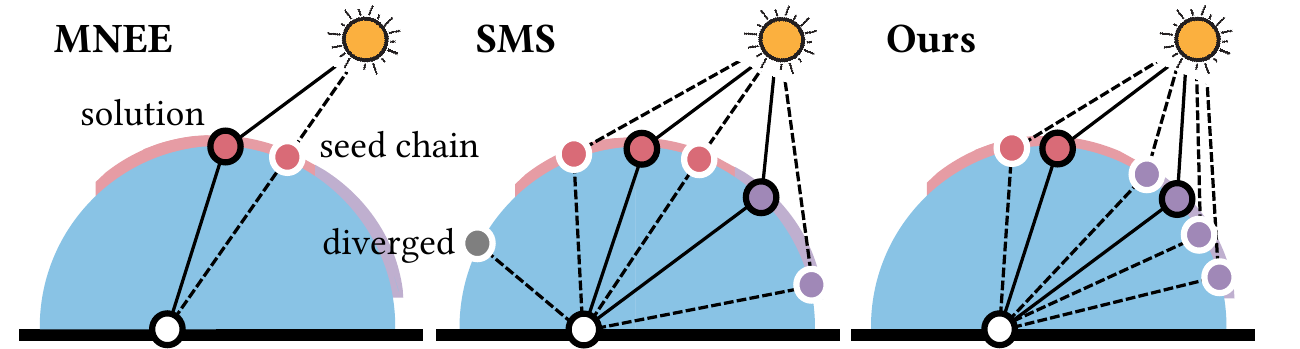}
	\end{minipage}
	\begin{minipage}{1.00\linewidth}
  			\includegraphics[width=1.0\linewidth]{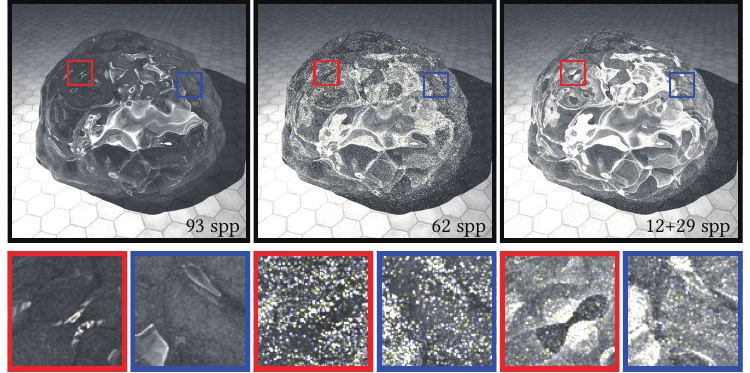}
	\end{minipage}
  \caption{
  MNEE \cite{Hanika15} easily loses energy due to incomplete solutions, while SMS \cite{Zeltner20} produces results with high variance. Our importance sampling method addresses these issues. The bottom row shows equal-time (3 min) rendering results of these methods.
  }
  \Description{...}
  \label{fig_result_mnee}
\end{figure}

As illustrated in Fig. \ref{fig_result_mnee}, MNEE \cite{Hanika15} uses a Dirac delta distribution 
as $p(\overline{\mathbf{x}}_S | \mathbf{x}_D, \mathbf{x}_L)$. 
When multiple solutions for a given configuration exist, MNEE becomes biased (or helps little to reduce variance when combined with a path tracer) \cite{Hanika15}. 
SMS \cite{Zeltner20} adopts a uniform seed chain distribution to enable unbiased sampling. Consequently, the variance can be very high since the probability density ignores the impact of the throughput. The situation becomes worse when the number of specular bounces increases, where the size of each convergence basin becomes extremely small while the probability of finding each chain is still proportional to the size of its convergence basin.
This inspires us to reduce the variance and support long specular chains by finding a better seed chain distribution $p(\overline{\mathbf{x}}_S | \mathbf{x}_D, \mathbf{x}_L)$ that is continuous and takes energy distributions into consideration.

\section{Problem Formulation}

To obtain a proper seed chain distribution mentioned above, we first formulate the problem. 
Evaluating the throughput summation using Monte Carlo techniques requires sampling the admissible chain $\overline{\mathbf{x}}_S^*$ in $\mathcal{P}_S^*$ for a given configuration $(\mathbf{x}_D, \mathbf{x}_L)$.
To achieve this goal, we require a sampling technique that generates random samples of discrete admissible chains with probability ${P(\overline{\mathbf{x}}_S^{*(i)} | \mathbf{x}_D, \mathbf{x}_L)}$, \ff{where $i$ denotes the index of samples used in Monte Carlo estimations. Then we resort to the following estimator:}
\begin{equation}
\begin{aligned}
\left \langle 
\sum_{\overline{\mathbf{x}}_S^* \in \mathcal{P}_S^*} T (\mathbf{x}_D,\overline{\mathbf{x}}_S^*, \mathbf{x}_L)
\right \rangle
=&
\frac 1 N \sum_{i=1}^N \frac {T (\mathbf{x}_D,\overline{\mathbf{x}}_S^{*(i)}, \mathbf{x}_L)} {P(\overline{\mathbf{x}}_S^{*(i)} | \mathbf{x}_D, \mathbf{x}_L)}
\label{Eq:ThroughputMC}
\end{aligned}
\textrm{.}
\end{equation} 
The estimation is unbiased if 
\begin{equation}
T (\mathbf{x}_D,\overline{\mathbf{x}}_S^{*(i)}, \mathbf{x}_L)>0 \Rightarrow P(\overline{\mathbf{x}}_S^{*(i)} | \mathbf{x}_D, \mathbf{x}_L)>0,
\end{equation}
but the value of the sampling probability has a dramatic impact on the variance. 
Ideally, if the probability of sampling each admissible chain is proportional to the throughput, i.e., 
\begin{equation}
P(\overline{\mathbf{x}}_S^{*(i)} | \mathbf{x}_D, \mathbf{x}_L) \propto T(\mathbf{x}_D,\overline{\mathbf{x}}_S^{*(i)}, \mathbf{x}_L),
\end{equation}
the estimation will reach zero variance and achieve the best possible importance sampling\footnote{The discussion here is under the assumption of diffuse separators, where the throughput equals the contribution to the full path.}.

Unfortunately, this task is not as simple as it seems and differs significantly from sampling non-specular vertices due to the discrete property of the $\mathcal{P}_S^*$ space.
Traditional path sampling techniques designed for continuous target distributions will fail since they hit each admissible chain with almost zero probability.

Our key insight is that, from a \textit{well-designed continuous distribution}, \ff{a \textit{seed chain}} can converge to
an admissible chain satisfying both the constraints and the requirements of importance sampling. 
Therefore, one key to the above problem is to find such a continuous distribution $p(\overline{\mathbf{x}}_S | \mathbf{x}_D, \mathbf{x}_L)$ from existing information about the discrete admissible chains.

\begin{figure}[tbp]
  \centering
  \includegraphics[width=1.0\linewidth]{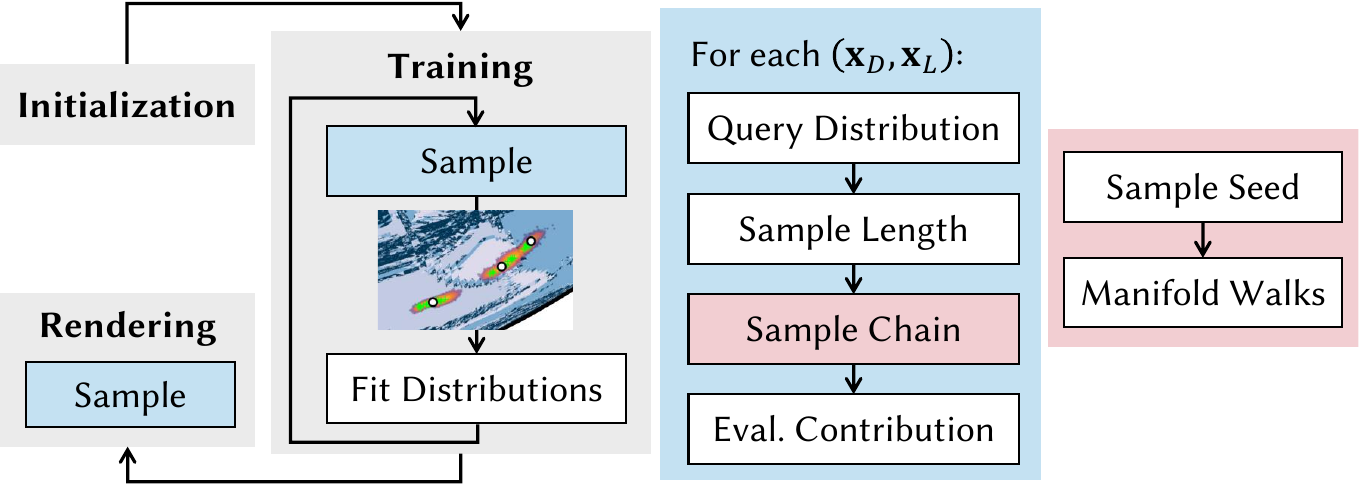} 
  \caption{\textbf{An overview of our proposed pipeline}. The pipeline employs a \ff{progressive manner} to gather sub-path samples (indicated by green plus signs), which are then utilized to fit continuous distributions (marked in orange). These distributions are essential for the importance sampling of seed chains, which ultimately converge to admissible chains (marked by white dots) if they fall within the convergence basins (colored in blue \ff{with varying brightness to distinguish different basins}). }
  \Description{...}
  \label{fig_pipeline}
\end{figure}

\section{Manifold Path Guiding}

In this section, we present a practical rendering method, \emph{manifold path guiding}, which allows us to address the above problem and hence achieve the goal of importance sampling specular chains.

To begin with, we will provide a brief summary of the pipeline, followed by discussions of each stage and some practical issues. Please refer to the supplemental document for pseudo-code snippets in Python style with explanations.

\subsection {Overview}
In general, as shown in Fig. \ref{fig_pipeline}, our pipeline consists of three stages: 
\begin{itemize}
    \item In the \textit{initialization} stage, we set an initial distribution that specifies how to sample seed chains without the knowledge of historical sub-path samples.
    \item During the \textit{training} stage, \ff{progressive training} is achieved by reconstructing distributions from previously found sub-path samples, sampling new seed chains, and performing manifold walk.
    \item In the \textit{rendering} stage, the final converged distribution is adopted to estimate the throughput between separators.
\end{itemize}
As aforementioned, we focus on specular chain sampling. Therefore, in our current implementation, we only handle the case in which $\mathbf{x}_L$ lies on light sources. We pick $\mathbf{x}_D$ by path tracing and $\mathbf{x}_L$ by uniformly sampling all emitters.

Our main goal is to reconstruct continuous distributions using the historical sub-path samples from previous iterations. After progressive refinement, the continuous distribution is expected to reflect the actual energy distribution and approach the optimal distribution for seed chain importance sampling. 

Therefore, the foundation of our online training and rendering pipeline is a seed chain importance sampling strategy that leverages the historical sub-path samples.
This is made possible by the following analyses of the continuity of the specular manifold and the light transport behaviors of specular paths.

\begin{figure}[tbp]
  \centering
  \includegraphics[width=1.0\linewidth]{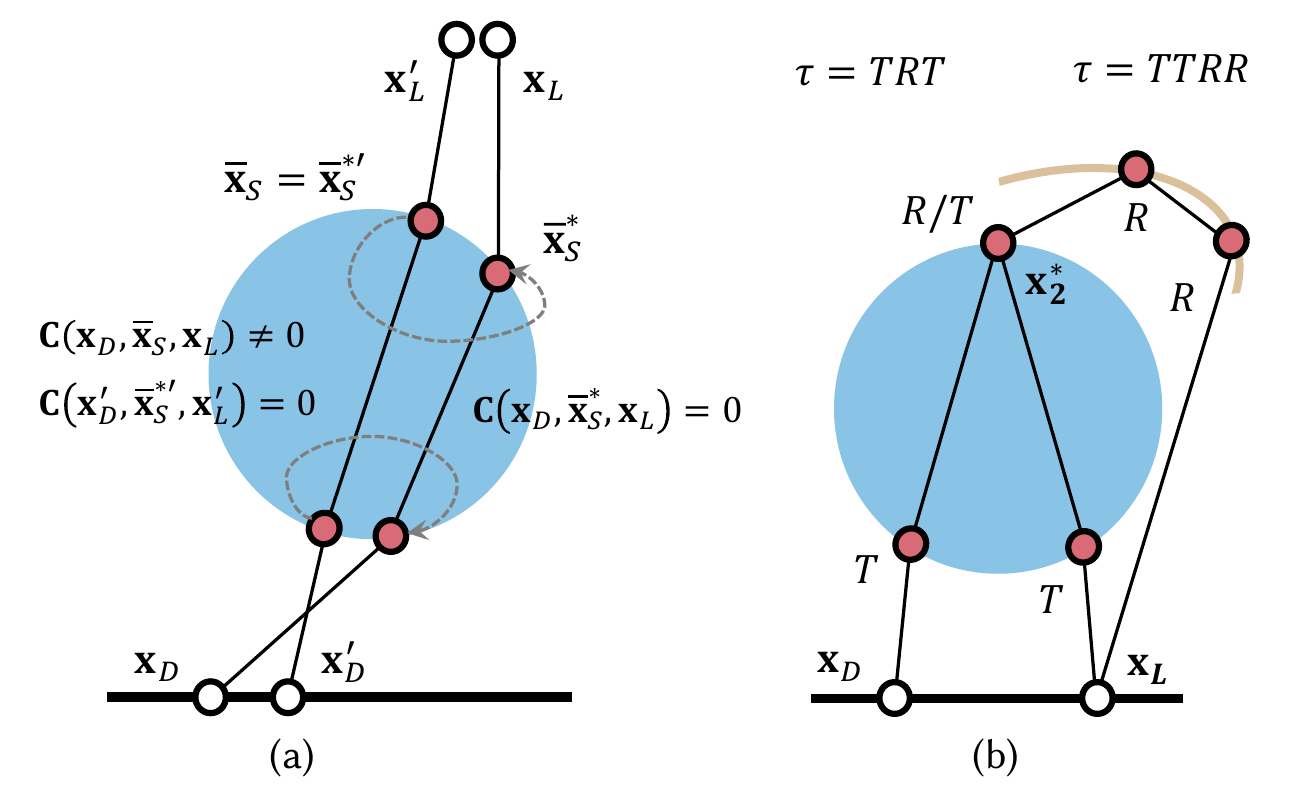}
  \caption{\textbf{(a)} Two configurations $(\mathbf{x}_D, \mathbf{x}_L)$ and $(\mathbf{x}_D', \mathbf{x}_L')$ that are close to each other. They tend to generate admissible sub-paths $(\mathbf{x}_D, \overline{\mathbf{x}}_S^*, \mathbf{x}_L)$ and $(\mathbf{x}_D', {\overline{\mathbf{x}}_S^*}', \mathbf{x}_L')$ through similar specular chains $\overline{\mathbf{x}}_S^*$ and ${\overline{\mathbf{x}}_S^*}'$. An admissible chain $\overline{\mathbf{x}}_S^*$ can be acquired starting from a seed chain $\overline{\mathbf{x}}_S = {\overline{\mathbf{x}}_S^*}'$ using manifold walk. \textbf{(b)}
  The scattering type is essential to ensure the uniqueness of the simplified chain description since reflection and refraction can happen at the same vertex ($\mathbf{x}_2^*$) simultaneously.
  }
  \Description{...}
  \label{fig_cor}
\end{figure}
\subsection{Exploitation of continuity}

Although the admissible chains under a specific configuration are usually discrete, all sub-paths in the scene form manifolds in the specular chain space $\mathcal{P}_S$, i.e., the specular manifolds \cite{Jakob12, Zeltner20}.

The continuity of the specular manifold is guaranteed by the Implicit Function Theorem \cite{Jakob12}. 
It tells that the function $\overline{\mathbf{x}}_S^*(\mathbf{x}_D, \mathbf{x}_L)$ is continuous in an infinitely small neighborhood around an admissible sub-path, and $\overline{\mathbf{x}}_S^*$ lies in the convergence basin\footnote{Manifold walks have quadratic convergence to an admissible chain when the seed is close to the solution \cite{Jakob12, Zeltner20}. This guarantees the existence of a convergence basin in the \textit{neighborhood} of an admissible chain, which can be used to solve for an admissible chain starting from an approximated one.} of that admissible sub-path. 
This allows us to use the admissible chain ${\overline{\mathbf{x}}_S^*}'$ of a nearby configuration $(\mathbf{x}_D', \mathbf{x}_L')$ to be the seed chain $\overline{\mathbf{x}}_S$ of the current configuration $(\mathbf{x}_D, \mathbf{x}_L)$. From this seed chain $\overline{\mathbf{x}}_S$, an admissible sub-path $(\mathbf{x}_D, \overline{\mathbf{x}}_S^*, \mathbf{x}_L)$ could be obtained after convergence. 
Fig. \ref{fig_cor}(a) gives a visual explanation of this process.

Moreover, the continuity of specular manifolds also results in the continuity of the sub-path throughput\footnote{In the neighborhood $\mathcal{N}$ of an admissible sub-path, $\overline{\mathbf{x}}_S^*$ has continuous partial derivatives to $\mathbf{x}_D$ and $\mathbf{x}_L$, which reveals the continuity of GGT. Assuming that $\kappa$ and $L$ in Eq. (\ref{Eq:ThroughputEval}) are locally continuous, the sub-path throughput $T$ is also continuous.}, which means the throughput is nearly constant in the neighborhood. Consequently, we can optimally importance sample the admissible chain under the current configuration by sampling the admissible chain under all nearby configurations according to their throughput. This will be particularly efficient when the throughput has a small value of derivatives to $\mathbf{x}_D$ and $\mathbf{x}_L$.
Now the problem has been significantly simplified --- the distribution of all nearby admissible chains is a continuous distribution, which can be fitted with any possible continuous distribution reconstruction technique.

\subsection{Dimensionality reduction}

As the length of the specular chain increases, there will be a risk of the curse of dimensionality. We avoid this by using a marginal. Let $\tau$ denote the \textit{type} of a specular chain $\overline {\mathbf{x}}_S$. It is a string consisting of $T$ and/or $R$, with $i$-th letter $\tau_i$ describing the scattering type at $i$-th vertex of the chain. For a specific configuration $(\mathbf{x}_D, \mathbf{x}_L)$, once $\tau$ is given and $\mathbf{x}_1 = \mathbf{r}(\mathbf{x}_D,\mbomega_D)$ is determined, the rest of the vertices $\mathbf{x}_2, ..., \mathbf{x}_n$ can be deduced using the following recursion:
\begin{equation}
\mathbf{x}_i=\mathbf{r}(\mathbf{x}_{i-1}, \mathbf{s}_{\tau_{i-1}}(\mathbf{x}_{i-1}, \overrightarrow{\mathbf{x}_{i-2}\mathbf{x}_{i-1}}))
\label{deduction}
\end{equation}
where the operator $\mathbf{r}(\mathbf{x},\mbomega )$ returns the intersection of the ray $(\mathbf{x},\mbomega)$ with all specular surfaces $\mathcal{M}_S$, and the operator $\mathbf{s}$ returns the scattered direction for given scattering type. Consequently, an admissible sub-path $(\mathbf{x}_D,\overline{\mathbf{x}}_S^*,\mathbf{x}_L)$ can be determined with $(\mathbf{x}_D, \mbomega_D^*, \mathbf{x}_L, \tau)$.
Note that the scattering type is necessary to ensure uniqueness since reflection and refraction can happen at the same vertex simultaneously, as visually explained in Fig. \ref{fig_cor}(b). 

Following the above representation, we can sample $(\mbomega_D,\tau)$ and use Eq. (\ref{deduction}) to obtain a seed chain $\overline{\mathbf{x}}_S$. It converts a high-dimensional sampling problem into a much lower one, and the probability  of sampling a seed chain can be further subdivided into two factors, i.e.,
\begin{equation}
p(\overline{\mathbf{x}}_S | \mathbf{x}_D,\mathbf{x}_L)
=
P(n | \mathbf{x}_D,\mathbf{x}_L)
p(\mbomega_D, \tau | \mathbf{x}_D,\mathbf{x}_L,n).
\label{eq_div1}
\end{equation}

With the above analysis, we divide the sampling task of a seed chain into three parts: sampling the number of bounces $n$, sampling the scattering type $\tau$, and sampling the direction $\mbomega_D$. This lays the foundation for efficient data-driven importance sampling. Note that the order of the last two steps may be changed due to practical considerations.

\subsection{Initialization}

The initial distribution $p_0(\overline{\mathbf{x}}_S|\mathbf{x}_D,\mathbf{x}_L)$ decides how initial seed chains are sampled only according to a priori knowledge. 
In theory, any distribution that covers all the admissible chains can be selected. Better choices can be adopted with the help of other existing algorithms. For instance, we can reconstruct energy distributions from photons tracing from emitters. Since reconstructed distributions are not fully conservative, we could further incorporate a uniform distribution to avoid making some regions never discovered \cite{Zhu2020PhotonDrivenNP, Muller17, Li22}.

In our implementation, we use the following initialization: 
\begin{itemize}
    \item The length of the chain is sampled by simulating a Russian Roulette (RR) defined in the underlying path tracer.
    \item The direction $\mbomega_D$ is determined by uniformly sampling positions as the first specular vertex $\mathbf{x}_1$ on all specular objects, yielding $p_0(\mbomega_D|\mathbf{x}_D,\mathbf{x}_L)$. 
    \item The remaining specular vertices $\mathbf{x}_2, ..., \mathbf{x}_n$ are generated by ray tracing, deciding the scattering type $\tau$ 
\end{itemize}
This covers all the possible $(\mbomega_D, \tau)$ pairs and is equivalent to covering all the admissible chains according to Eq. (\ref{eq_div1}). Consequently, each solution will be reached with non-zero probability.

\subsection{Importance sampling seed chains}

The continuity discussed in Section 5.2 allows us to perform importance sampling leveraging historical samples of neighboring configurations. 
Now we reconstruct discrete distributions for $n$ and $\tau$ and continuous distributions for $\mbomega_D$, rewriting Eq. (\ref{eq_div1}) as
\begin{equation}
p(\overline{\mathbf{x}}_S | \mathbf{x}_D,\mathbf{x}_L)
=
P(n | \mathbf{x}_D,\mathbf{x}_L)
P(\tau | \mathbf{x}_D,\mathbf{x}_L, n)
p(\mbomega_D | \mathbf{x}_D,\mathbf{x}_L,\tau)
\mathrm{.}
\label{eq_div2}
\end{equation}

\paragraph{Gathering neighboring samples}

We first perform a neighbor searching process in a set\footnote{$\mathcal{S}$ can include either all the samples already encountered \cite{Reibold2018SelectiveGS} or only the samples found in the last iteration \cite{Muller17}. We use the latter by default.} of previously obtained sub-path samples $\mathcal{S}$, gathering a set of sub-paths with the nearest configurations to $(\mathbf{x}_D,\mathbf{x}_L)$, denoted as $\mathcal{S}(\mathbf{x}_D,\mathbf{x}_L)$. 
\ff{Then, we evaluate the weight of a sub-path sample, $w(\mathbf{x}_D', {\overline{\mathbf{x}}_S^*}', \mathbf{x}_L')$, which equals to the product of its throughput and the reciprocal probability estimation:
\begin{equation}\label{eq_weight}
w(\mathbf{x}_D', {\overline{\mathbf{x}}_S^*}', \mathbf{x}_L') =T(\mathbf{x}_D', {\overline{\mathbf{x}}_S^*}', \mathbf{x}_L')
\left \langle \frac{1}{P({\overline{\mathbf{x}}_S^*}' | \mathbf{x}_D', \mathbf{x}_L')} \right \rangle.
\end{equation}
Here, the throughput and reciprocal probability are computed in earlier iterations and stored in memory.}

\paragraph{Sampling chain length}

The length of a specular chain determines the dimensionality of the specular chain space and thus needs to be sampled first. 
By accumulating and normalizing the weights for each length, we obtain a discrete probability distribution $P_e(n|\mathbf{x}_D,\mathbf{x}_L)$\footnote{\ff{We use subscript $e$ for fitted distributions and subscript $0$ for initial distributions.}} from $\mathcal{S}(\mathbf{x}_D,\mathbf{x}_L)$:
\ff{\begin{equation}
P_e(n | \mathbf{x}_D,\mathbf{x}_L)
=
\frac{
\sum_{(\mathbf{x}_D', {\overline{\mathbf{x}}_S^*}', \mathbf{x}_L') \in \mathcal{S}(\mathbf{x}_D,\mathbf{x}_L,n)} 
w(\mathbf{x}_D', {\overline{\mathbf{x}}_S^*}', \mathbf{x}_L')
}{
\sum_{(\mathbf{x}_D', {\overline{\mathbf{x}}_S^*}', \mathbf{x}_L') \in \mathcal{S}( \mathbf{x}_D,\mathbf{x}_L)} 
w(\mathbf{x}_D', {\overline{\mathbf{x}}_S^*}', \mathbf{x}_L')
},
\end{equation}}
where $\mathcal{S}(\mathbf{x}_D, \mathbf{x}_L,n)$ denotes all the samples of length $n$ in $\mathcal{S}(\mathbf{x}_D, \mathbf{x}_L)$. 
This will be used when sampling the length of a new seed chain.

\paragraph{Sampling scattering type}

Once $n$ is determined, we sample the scattering type from a discrete distribution reconstructed by summing up the total weights of each type:
\ff{\begin{equation}
P_e(\tau | \mathbf{x}_D,\mathbf{x}_L, n)
=
\frac{
\sum_{(\mathbf{x}_D', {\overline{\mathbf{x}}_S^*}', \mathbf{x}_L') \in \mathcal{S}(\mathbf{x}_D,\mathbf{x}_L,\tau)} 
w(\mathbf{x}_D', {\overline{\mathbf{x}}_S^*}', \mathbf{x}_L')
}{
\sum_{(\mathbf{x}_D', {\overline{\mathbf{x}}_S^*}', \mathbf{x}_L') \in \mathcal{S}(\mathbf{x}_D,\mathbf{x}_L,n)} 
w(\mathbf{x}_D', {\overline{\mathbf{x}}_S^*}', \mathbf{x}_L')
}.
\end{equation}}
Let $\mathcal{S}(\mathbf{x}_D,\mathbf{x}_L,\tau)$ denote the set of samples with type $\tau$. The final step is sampling the direction from $\mathcal{S}(\mathbf{x}_D,\mathbf{x}_L,\tau)$.

\paragraph{Sampling the direction}

We blur each chain in $\mathcal{S}(\mathbf{x}_D,\mathbf{x}_L,\tau)$ with a wide kernel in the directional domain.
Specifically, we estimate the directional footprint \cite{Hey02} of $i$-th chain by finding a sample in the $\mathcal{S}(\mathbf{x}_D,\mathbf{x}_L,\tau)$ with the nearest \ff{${\mbomega_D^*}'$} to it and evaluating their directional distance \ff{$\sigma_i'$}. 
Then, we place a von Mises-Fisher (vMF) lobe \cite{Fisher53} along the direction ${\mbomega_D^*}_i'$:
\begin{equation}
v(\mbomega_D; \mbmu_i, \kappa_i) 
=
\frac {\kappa_i} {4 \pi \sinh(\kappa_i)} e^{\kappa_i \mbmu_i \cdot \mbomega_D}
\label{Eq:vMF}
\end{equation}
where we set \ff{$\mbmu_i={\mbomega_D^*}_i'$, $\kappa_i = {\sigma_i'}^{-2}$}.
With the weights of each lobe being proportional to the corresponding sub-path sample's weight defined in Eq. (\ref{eq_weight}), we obtain a mixture of vMF lobes $p_e(\mbomega_D | \mathbf{x}_D, \mathbf{x}_L, \tau)$.

\begin{figure}[tbp]
  \centering
  \includegraphics[width=1\linewidth]{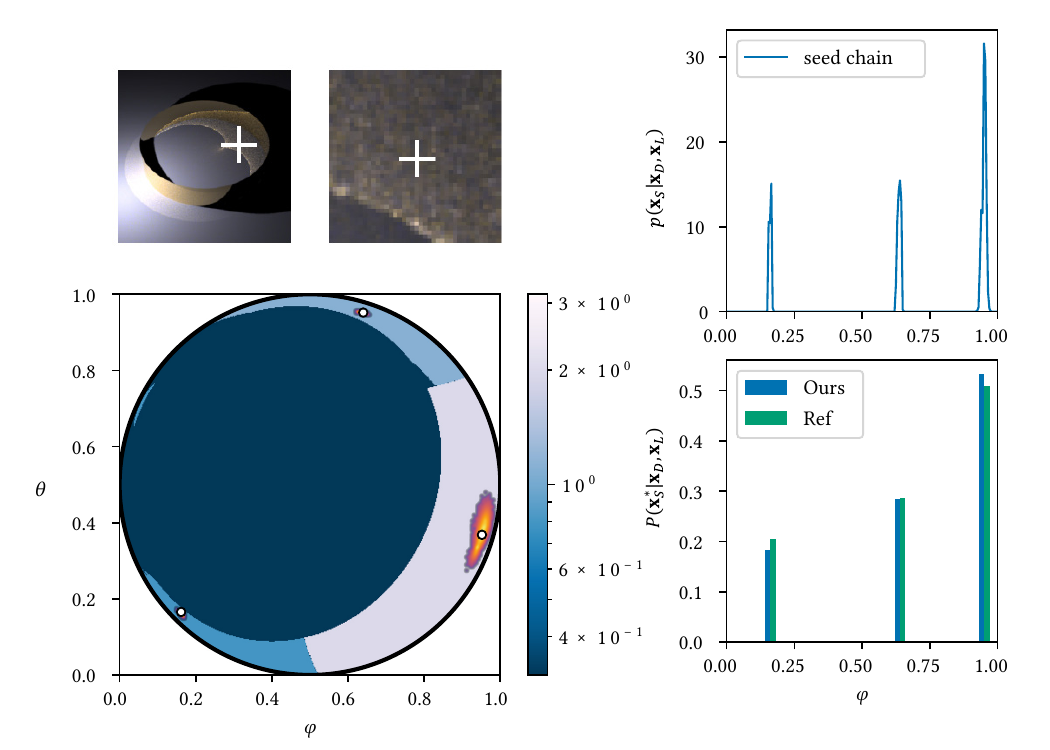} 
  \caption{\textbf{An example validating our solution of importance sampling discrete specular chains.} We choose a typical configuration in the \textsc{Ring} scene: $\mathbf{x}_L$ on a small emitter and $\mathbf{x}_D$ marked by the white plus sign. The convergence basin and our sampling distribution are visualized in spherical coordinates. Two diagrams in the right panel visualize the discrete distribution of admissible chains ${P(\overline{\mathbf{x}}_S^{*(i)} | \mathbf{x}_D, \mathbf{x}_L)}$ for this configuration (bottom-right) and the continuous distribution $p(\overline{\mathbf{x}}_S | \mathbf{x}_D, \mathbf{x}_L)$ reconstructed from historical samples (top-right).}
  \Description{...}
  \label{fig_importance}
\end{figure}

\paragraph{Validation and discussion}

In Fig. \ref{fig_importance}, we verify that the proposed method can achieve nearly optimal importance sampling for discrete admissible chains. Here, we marginalize the reconstructed distribution of seed chains along the horizontal flatland, shown in the top-right diagram. In the bottom-right diagram, we evaluate the corresponding sampling probability (blue) and the expected probability according to their throughput (green). As seen, the probability of finding each admissible chain closely matches their throughput.
In theory, this is still possible with high-frequency details and complex visibility, despite requiring more samples.

\paragraph{\ff{Defensive} sampling.} 
In order to cover all admissible chains, we further perform Multiple Importance Sampling (MIS) \cite{Veach95MIS} with the initial distribution.
We use a one-sample MIS model for sampling both $n$ and $(\mbomega_D, \tau)$:
\begin{equation}
\begin{aligned}
P(n|\mathbf{x}_D,\mathbf{x}_L) &= \alpha P_0(n|\mathbf{x}_D,\mathbf{x}_L) + (1 - \alpha) P_e(n|\mathbf{x}_D,\mathbf{x}_L) \\
p(\mbomega_D, \tau|\mathbf{x}_D,\mathbf{x}_L) &= 
\alpha p_0(\mbomega_D, \tau|\mathbf{x}_D,\mathbf{x}_L) + 
(1 - \alpha) p_e(\mbomega_D, \tau|\mathbf{x}_D,\mathbf{x}_L) 
\end{aligned}
\end{equation}
where $\alpha$ is the probability of choosing $P_0$. Following the convention in many path guiding methods, we always set $\alpha=0.5$ \cite{Muller17, Zhu2021HierarchicalNR, Reibold2018SelectiveGS}.

\subsection{Spatial structures}

In Section 5.5, we gather nearby sub-path samples and construct vMF lobes to represent the directional distribution.
However, the choice of spatial hierarchy 
significantly affects efficiency and robustness.

Performing an accurate k-nearest neighbor search (implemented with a 6D kd-tree) would be computationally expensive. Instead, we use an approximated solution. We organize all the samples in a 6D adaptive binary tree (STree) following \cite{Muller17} and set the spatial threshold (i.e., the maximum number of samples in a leaf node) to $\sqrt{|\mathcal{S}|}$. This is generally similar to \cite{mueller19guiding}, but we use the number of total samples rather than the samples per pixel and omit the coefficient. Our strategy is independent of image resolution and works well in all our test scenes\footnote{One can further fine-tune the performance by using $c\sqrt{|\mathcal{S}|}$, as shown in our supplemental document. Generally, we recommend $c=1$, which is simple and straightforward.}. 
When subdividing nodes, we perform spatial filtering by allowing a small overlapping between child nodes\footnote{
Specifically, we copy $\varepsilon k$ sub-path samples from the left subtree to the right subtree and vice versa. We also discard the samples far from the bounding box of a node.
Overall, we observe that $\varepsilon=10\%$ works well across all our test scenes.
}.  
This prevents artifacts introduced by spatial division \cite{mueller19guiding, Zhu2021HierarchicalNR, Ruppert2020RobustFO}.

\begin{figure}[tbp]
\centering
	\begin{minipage}{1.00\linewidth}
  			\includegraphics[width=1.0\linewidth]{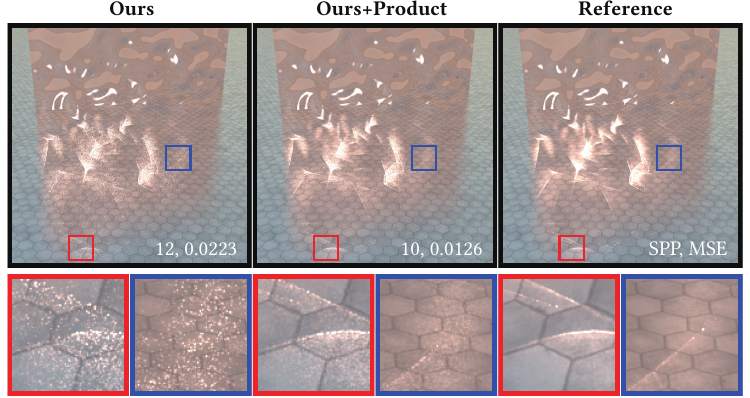} % finalized
	\end{minipage}
  \caption{\textbf{Equal-time (5 min) comparison} of our approach without and with product importance sampling with the separator's BSDF. A normal-mapped reflective specular plane is placed on a glossy ($\alpha = 0.05$) floor.} 
  \Description{...}
  \label{figure_result_product}
\end{figure}

\subsection{Handling glossy cases}

Our earlier discussions are under the assumption of diffuse separators and pure specular chains. However, our method can be generalized to support glossy separators and chains.

\paragraph{Glossy separators.}

When the separator is not diffuse, 
to achieve optimal importance sampling, we should 
perform product importance sampling \cite{Herholz2016ProductIS} with the separator's BSDF. 
Since we use sub-path samples to reconstruct the distributions, we can directly modify their weights before sampling, and \ff{Eq. (\ref{eq_weight}) becomes
\begin{equation}\label{eq_weight_pis}
w(\mathbf{x}_D', {\overline{\mathbf{x}}_S^*}', \mathbf{x}_L') = \rho_{\mathbf{x}_D'}(\mbomega_i', {\mbomega_D^*}') T(\mathbf{x}_D', {\overline{\mathbf{x}}_S^*}', \mathbf{x}_L')
\left \langle \frac{1}{P({\overline{\mathbf{x}}_S^*}' | \mathbf{x}_D', \mathbf{x}_L')} \right \rangle.
\end{equation}
Here, $\mbomega_i'$ in the BSDF term $\rho_{\mathbf{x}_D'}(\mbomega_i', {\mbomega_D^*}')$ is the direction towards $\mathbf{x}_D'$} and is already determined. 
All we need is to traverse through each sub-path sample in $\mathcal{S}(\mathbf{x}_D,\mathbf{x}_L)$ and apply the above weighting scheme.
This can significantly reduce noise when separators have small roughness, as shown in Fig. \ref{figure_result_product}. 

\paragraph{Glossy chains}

For glossy chains, once manifold offset \cite{Jakob12, Kaplanyan2014} is sampled, it boils down to pure specular cases, and the admissible chains corresponding to the offset are still finite. This is unbiased since each chain can be found with non-zero probability. Like prior work \cite{Kaplanyan2014,Hanika15,Zeltner20}, we sample the \textit{offset normal} from the microfacet distribution before manifold sampling. Then, we sample seed chains and admissible chains constrained to this offset normal.

\subsection{Practical considerations}

\begin{figure}[tbp]
\centering
	\begin{minipage}{1.00\linewidth}
  			\includegraphics[width=1\linewidth]{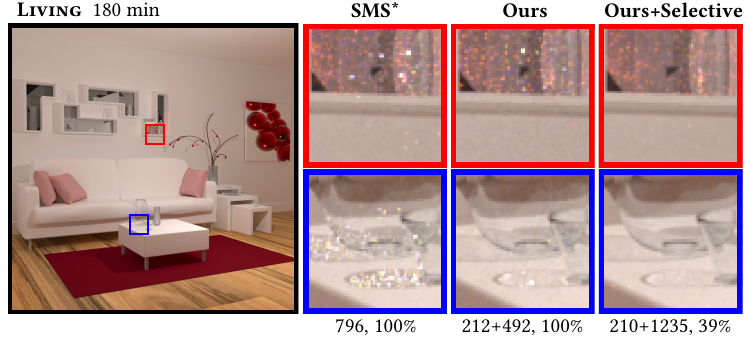} % finalized
  			
	\end{minipage}
  \caption{\textbf{Selective activation} on a realistic scene, where caustics only cover a small part of images. We report SPP and the activation rates.  }
  \Description{...}
  \label{fig_result_selective}
\end{figure}

\paragraph{Reciprocal probability estimation.}
To estimate the reciprocal probability in Eq. (\ref{eq_weight}), we repeat independent trials once the sampled admissible chain was found, in the same way as in SMS \cite{Zeltner20}. However, some portions of the reciprocal probability can be evaluated analytically. In particular, \ff{$P(n|\mathbf{x}_D,\mathbf{x}_L)$} is determined when we sample the length of the chain. Therefore, it can be factored out of the stochastic estimation, i.e., 
\begin{equation}
\left \langle \frac{1}{P(\overline{\mathbf{x}}_S^* | \mathbf{x}_D, \mathbf{x}_L)} \right \rangle
=
\frac{1}{
P(n|\mathbf{x}_D,\mathbf{x}_L)}
\left \langle \frac{1}{P(\overline {\mathbf{x}}_S^*|\mathbf{x}_D,\mathbf{x}_L,n)} \right \rangle
\end{equation}
and only $P(\overline {\mathbf{x}}_S^*|\mathbf{x}_D,\mathbf{x}_L,n)$ needs to be estimated in our implementation. In other words, we \textit{reuse} $n$ across trials.

\paragraph {Online \ff{training} and rendering}

We use an online \ff{training} strategy that progressively gathers sub-path samples. We separate the training stage into several iterations with increasing budgets. Specifically, we double the sample count in each iteration \cite{Muller17}. We stop the training stage immediately when 30\% of the budget is already used. 
Besides, when a training iteration is finished, we start a new round if less than half of the training budget is used. Otherwise, we continue the current iteration until the rendering stage starts.
Considering that high variance may occur during the \ff{training} phase, we never splat the samples produced in the training stage into the final image. Besides, we only use samples from the last iteration to reconstruct sampling distributions.

\paragraph{Selective activation.}
Our learned distribution can also be employed to determine when specular chain sampling should be activated \cite{Loubet20}, which avoids the costly sampling process for non-caustic regions \cite{Zeltner20}.
In the rendering stage, if no samples (excluding those used for filtering) are in the corresponding leaf node of the spatial hierarchy for a given configuration, we simply switch to using path tracing. 
In other words, we perform a combination with path tracing according to the existence of nearby sub-path samples.
This significantly reduces overhead and leads to more samples in equal time, as illustrated in Fig. \ref{fig_result_selective}.

\begin{figure*}[htbp]
\centering
	\begin{minipage}{1.00\linewidth}
  			\includegraphics[width=1.0\linewidth]{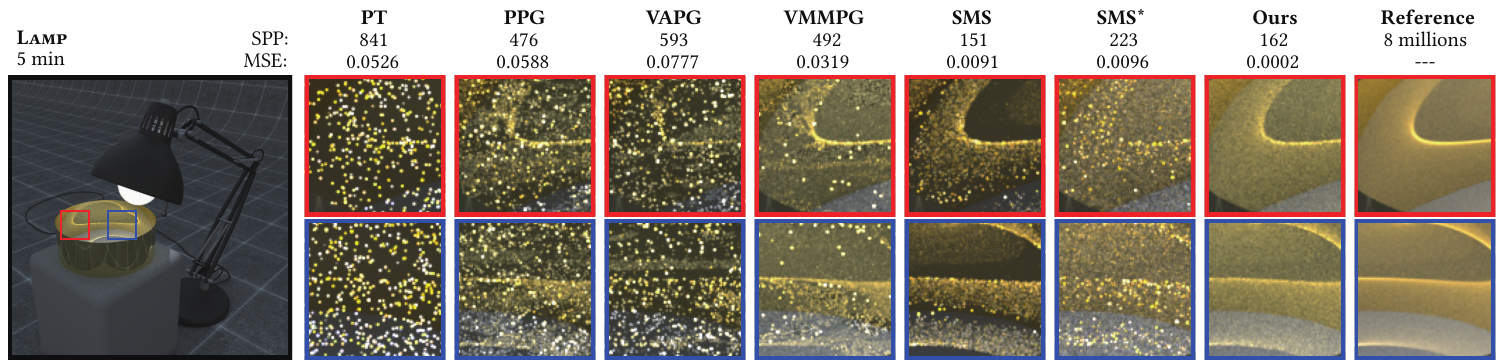}
	\end{minipage}
	\begin{minipage}{1.00\linewidth}
  			\includegraphics[width=1.0\linewidth]{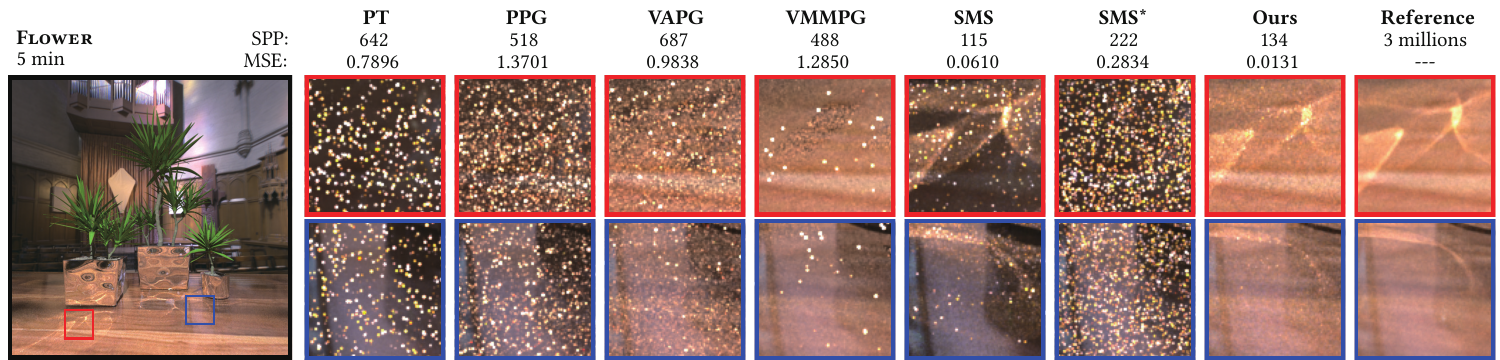}
	\end{minipage}
	\begin{minipage}{1.00\linewidth}
  			\includegraphics[width=1.0\linewidth]{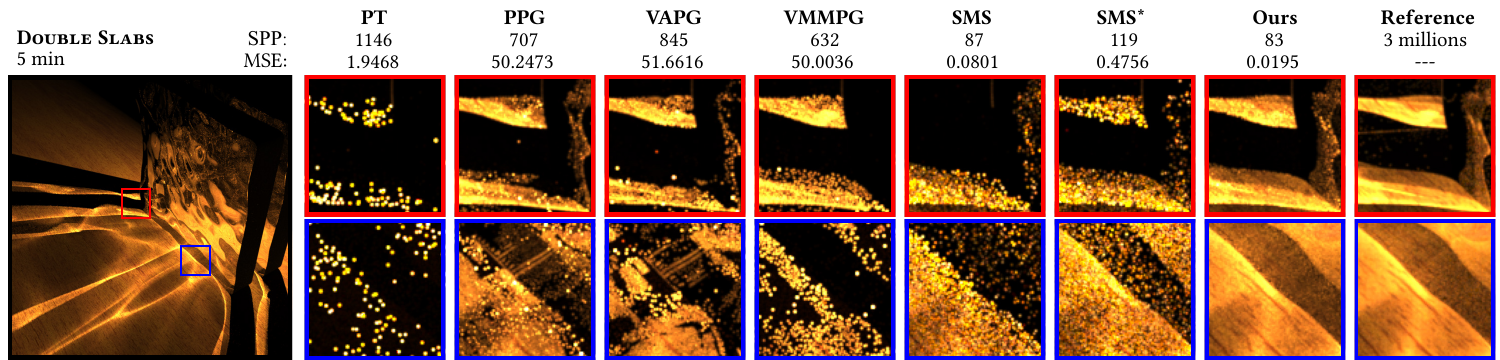} % finalized
	\end{minipage}
  \caption{\textbf{Equal-time comparison\ff{s}} among Path Tracing (PT), the improved Practical Path Guiding (PPG) \cite{Muller17, mueller19guiding}, Variance-Aware Path Guiding (VAPG) \cite{Rath2020VarianceawarePG}, parallax-aware VMM Path Guiding (VMMPG) \cite{Ruppert2020RobustFO}, the original Specular Manifold Sampling (SMS) \cite{Zeltner20}, a modified version of SMS (SMS*) and our method on three scenes. 
  }
  \Description{...}
  \label{fig_result_2}
\end{figure*}

\section{Results}

We have implemented our approach in the Mitsuba 2 Renderer \cite{Mitsuba2} as a new integrator. 
In this section, we will describe our experimental setup, compare our approach to related methods, and validate our building blocks.

All our test scenes have been rendered on a compute node with 32 2.50 GHz cores of Xeon E5-2682 v4 processors. Since we focus on high-frequency light transport, we only use artificial point and small area light sources to cast caustics and handle them using the proposed approach. Other light transport portions involving environmental lighting are handled by conventional path tracing. We set the max length of paths to 15. Russian roulette starts from bounce 5 with probability $\gamma = 0.95$. Unless otherwise mentioned, we do \emph{not} enable product importance sampling or selective activation for a more fair comparison with related methods.

The reference images of \textsc{Lamp}, \textsc{Flower}, and \textsc{Double Slabs} are generated using path tracing, whereas the others are rendered with our modified SMS at high sample rates. \ff{Some of our results report the number of samples per pixel in the form of ``$n+m$'', meaning $n$ spp for training and $m$ spp for rendering.}

\subsection{Comparisons with previous approaches}

\paragraph{Comparison with unbiased MC approaches}

In Fig. \ref{fig_result_2}, we show the equal-time comparison of our approach against PT, PPG \cite{mueller19guiding}, VAPG \cite{Rath2020VarianceawarePG}, VMMPG \cite{Ruppert2020RobustFO}, the original SMS \cite{Zeltner20}, and our modified SMS on three scenes: \textsc{Lamp}, \textsc{Flower} and \textsc{Double Slabs}.

\begin{figure}[tbp]
\centering
	\begin{minipage}{1.00\linewidth}
  			\includegraphics[width=1.0\linewidth] {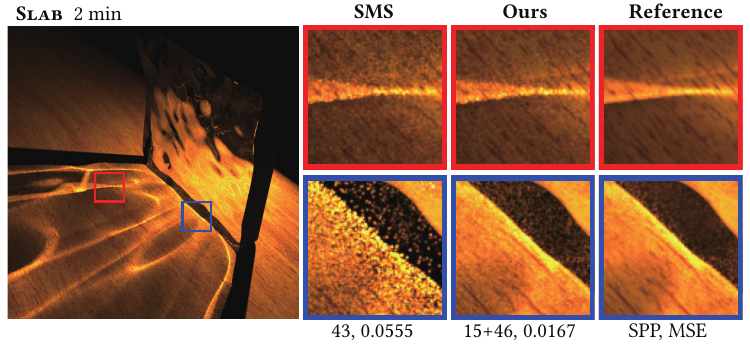} % finalized
	\end{minipage}
  \caption{\textbf{Equal-time comparison} between the original SMS \cite{Zeltner20} and our method on the \textsc{Slab} scene. In order to validate the effect of directional sampling, we use a fixed chain length ($n=2$) and also neglect reflection on the dielectric surface. 
  }
  \Description{...}
  \label{fig_result_1}
\end{figure}  
\begin{figure}[tbp]
\centering
	\begin{minipage}{1.00\linewidth}
  			\includegraphics[width=0.5\linewidth]{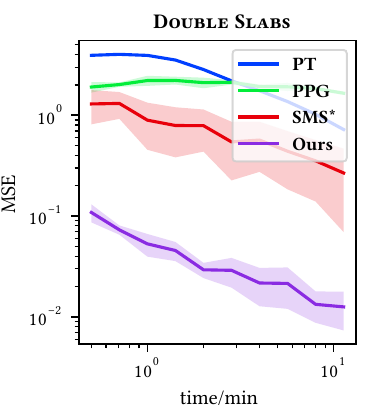} % finalized
  			\includegraphics[width=0.5\linewidth]{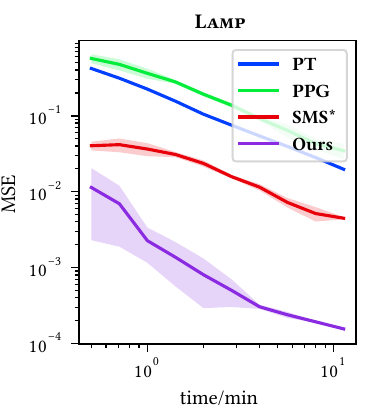} % finalized
	\end{minipage}
\caption{\textbf{Convergence curves of two testing scenes.} 
We compare our method with PT, PPG, and SMS* on the evolution of MSE with respect to the total rendering (including training) time. The standard deviation of the MSE is shown as shaded regions.
} 
  \Description{...} 
  \label{fig_result_convergence}
\end{figure}
Conventional guiding approaches, such as PPG, rely on online learned distributions to perform path sampling. They generally work well but will encounter difficulties in dealing with glossy or specular interactions since the radiance distribution involves high-frequency variations. Visual comparisons clearly show that a typically guided path tracer, such as PPG, VAPG, and VMMPG, fails to handle challenging light paths involving specular vertices, leading to many outliers and energy loss. 

Although our approach also adopts reconstructed energy distributions as guidance, its usage is quite different. First, our distribution is reconstructed from admissible sub-path samples, which is beneficial for exploring high-frequency regions. Second, the reconstructed distribution is used to sample seed chains instead of light paths. Therefore, even inaccurate distributions can still work well. Consequently, our method succeeds in covering all challenging lighting effects with much fewer fireflies.

\begin{figure}[tbp]
\centering
  			\includegraphics[width=1\linewidth]{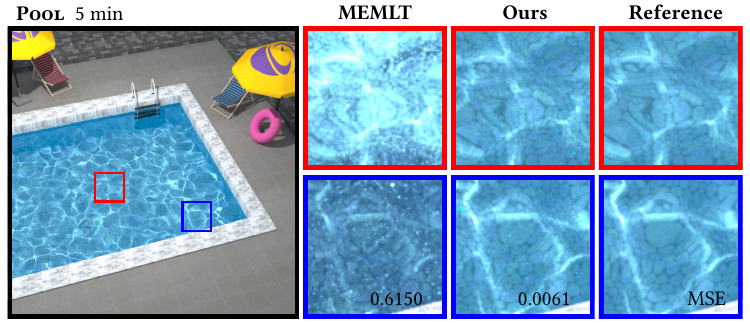}
     
     \includegraphics[width=1\linewidth]{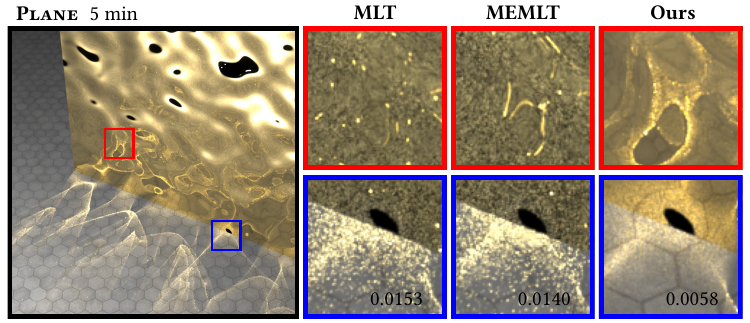}
  			
  \caption{
  \textbf{Equal-time (5 min) comparison} between MLT \cite{Veach97MLT}, MEMLT \cite{Jakob12} and our method.
  } 
  \Description{...}
  \label{fig_result_other}
\end{figure}

SMS is the state-of-the-art approach tailored for the unbiased simulation of specular light transport. The original SMS can only support a user-specified and fixed chain length. We extend it (denoted as SMS*) using our initialization (Section 5.4) to support various specular chain lengths and scattering types. 
Although this enables handling all types of specular chains, it has the risk of higher variance (see the \textsc{Lamp} scene) than the original SMS since the sample budget is amortized for various types of chains.

Thus, to further demonstrate the effect of our importance sampling strategy, we also choose a dominant type of chain for each scene ($RT$ for \textsc{Lamp}, $TTR$ for \textsc{Flower}, and $TTTT$ for \textsc{Double Slabs}) and present the results rendered by the original SMS in Fig. \ref{fig_result_2}. 
Even in this way, the original SMS still tends to produce noticeable noise since it samples seed chains uniformly.
This is particularly obvious in areas that correspond to admissible chains with extremely small convergence basins, e.g., the bottom of the slabs. 

Our method exploits progressively generated sub-path samples to reconstruct and refine continuous distributions for sampling seed chains, thus providing high-quality, unbiased rendering for challenging SDS paths. Even at relatively low sample rates, our method still outperforms its competitors with much lower variance.

We further validate the benefit of our directional importance sampling in Fig. \ref{fig_result_1}, where we disable the sampling of chain length and type in our method and compare with the original SMS to validate the effect of directional importance sampling. 
As highlighted in the closeups, due to efficient importance sampling, our approach generates more samples and produces results with far less noise. 
A complete ablation study is shown in the supplemental document, which validates the effectiveness of length sampling, type sampling, and directional sampling, respectively.

In Fig. \ref{fig_result_convergence}, we evaluate the accuracy in terms of MSE with respect to the running time, showing the fast convergence provided by our method. Thanks to the energy-based sampling distribution, our method consistently outperforms previous ones as the running time increases and achieves a large margin at a high sampling rate.

\begin{figure}[tbp]
\centering
  			
  			\includegraphics[width=1.0\linewidth]{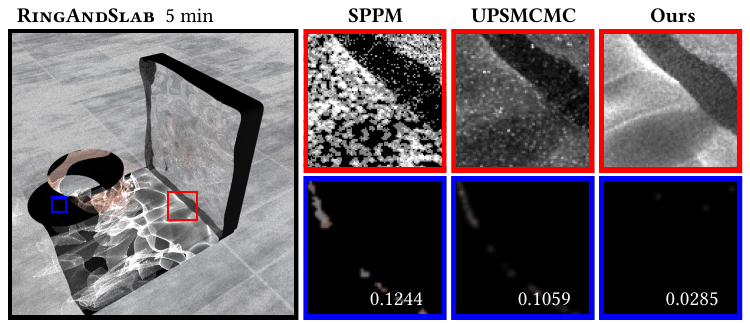}
  \caption{
  \textbf{Equal-time (5 min) comparison} between SPPM \cite{Hachisuka09SPPM}, UPSMCMC \cite{Sik2016RobustLT}, and our method. 
  } 
  \Description{...}
  \label{fig_result_biased}
\end{figure}

\paragraph{Comparison with unbiased MCMC approaches} 

\begin{figure*}[tbp]
\centering
 
	\begin{minipage}{0.495\linewidth}
  		\includegraphics[width=1.0\linewidth]{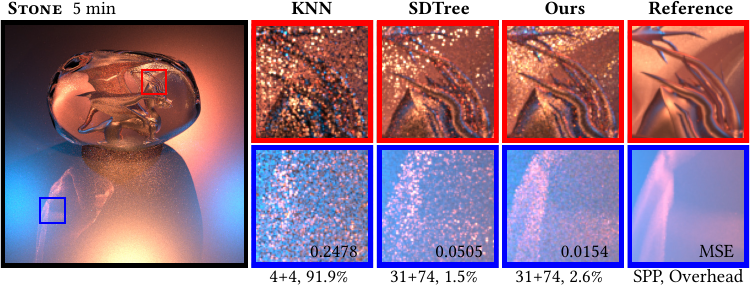} 
	\end{minipage}
	\begin{minipage}{0.495\linewidth}
  		\includegraphics[width=1.0\linewidth]{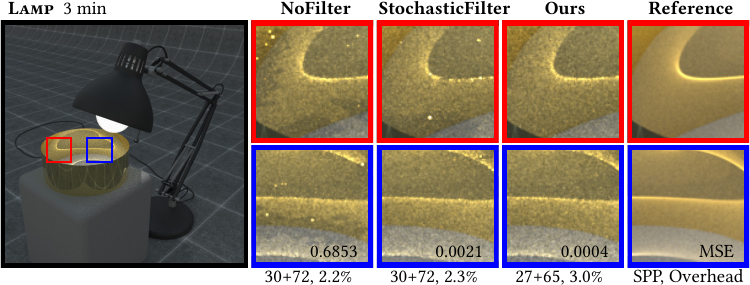} 
	\end{minipage}

  \caption{\textbf{Choices of building blocks.}
  \textbf{Left:}
  we perform equal-time comparisons of various spatial and directional structures. 
  \textbf{Right:}
  comparisons of different spatial filtering strategies.
  We report the full-image MSE, SPP, and the overhead of sampling. 
  More scenes are shown in the supplemental document.
  } 
  \Description{...}
  \label{fig_val_spatial}
\end{figure*}

\begin{figure*}[tbp]
\centering

	\begin{minipage}{0.33\linewidth}
  		\includegraphics[width=1.0\linewidth]{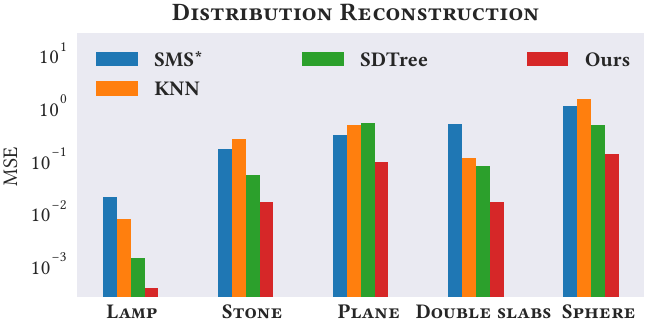} 
	\end{minipage}
	\begin{minipage}{0.33\linewidth}
  		\includegraphics[width=1.0\linewidth]{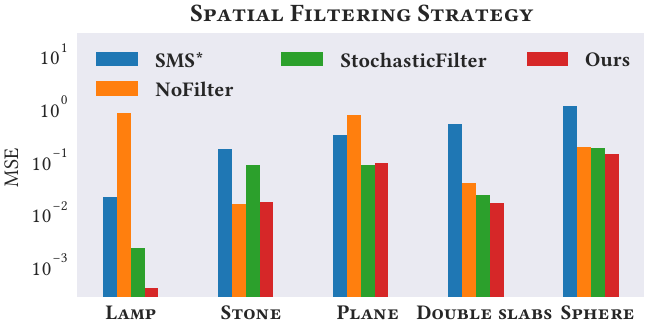} 
	\end{minipage}
	\begin{minipage}{0.33\linewidth}
  		\includegraphics[width=1.0\linewidth]{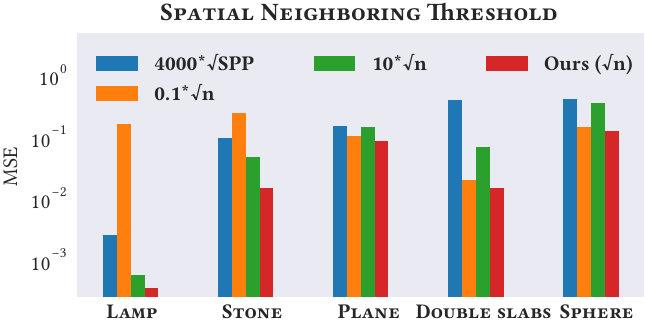} 
	\end{minipage}
  \caption{\textbf{Equal-time comparison of different distribution reconstruction techniques, spatial filtering strategies, and neighboring thresholds }on five scenes featuring different light transport effects and complexity. The corresponding images are shown in the supplemental document. 
  } 
  \Description{...}
  \label{fig_val_spatial_bar}
\end{figure*}

In Fig. \ref{fig_result_other}, we compare our approach with MLT \cite{Veach97MLT} and MEMLT \cite{Jakob12}, two typical MCMC approaches. The 
\textsc{Pool} scene is a typical example of challenging paths, and the 
\textsc{Plane} scene contains a normal-mapped conductor with roughness ($\alpha=0.002$).
Both MLT and MEMLT require admissible chains as their seed paths. Simply relying on PT or BDPT to find valid seed paths is inefficient for SDS paths. This results in either energy loss if no seed path is found for a specific region (blue closeups in \textsc{Pool}) or over-bright artifacts where Markov chains get stuck in narrow regions (red closeups in both scenes). 
As our method works in the context of regular Monte Carlo sampling, it is able to produce images free from blotchy artifacts.

\paragraph{Comparison with biased approaches}

Biased approaches excel at producing noise-free results for complex caustics and hence prevail in the industry.
In Fig. \ref{fig_result_biased}, we conduct an equal-time comparison with Stochastic Progressive Photon Mapping (SPPM) \cite{Hachisuka09SPPM} and Metropolised Bidirectional Estimator (UPSMCMC) \cite{Sik2016RobustLT}.
A slab and a ring are lit by a distant point light encapsulated in a glass sphere.

For a meaningful comparison with SPPM, we decrease the photon lookup radius to the point that the bias is less perceptible. Despite this, the noise level is still higher than ours. It also suffers from visible light leaking at the bottom of the ring. A similar issue occurs for UPSMCMC.
Its MCMC nature also results in an uneven convergence characterized by an overly dark region and visible fireflies.

As an unbiased method, our technique performs independent Monte Carlo estimation without spatial relaxing or reuse, thereby avoiding these problems and producing caustics of high quality with sharp details and low noise.

Fig. \ref{fig_result_mnee} shows a visual comparison with MNEE \cite{Hanika15}. As aforementioned, MNEE cannot find all valid admissible chains in this scene since it relies on deterministic initialization. Generally, this method performs well in simple cases but works inefficiently and tends to be biased (or suffers from extremely high variance if combined with path tracing) in complex scenes with many solutions. The stochastic sampling strategy adopted in our method makes it possible to cover all solutions for a configuration and hence ensures unbiasedness.

\subsection{Validation of building blocks}

In theory, our pipeline supports most feasible spatial neighbor searching and directional density estimation techniques. However, the decision will significantly affect the quality of the sampling. Here, we endeavor to generate such insights by comparing various alternative choices of pipeline components in Fig. \ref{fig_val_spatial} and Fig. \ref{fig_val_spatial_bar}.

\paragraph{Distribution reconstruction.}

We employ STree for approximated spatial nearest neighbor searching when reconstructing the energy distribution, and perform an accurate directional density estimation. 
To evaluate their effectiveness, we compare our method against adaptive quad-tree \cite{Muller17} (\textit{SDTree}) and a variant of our approach with accurate spatial nearest neighbor searching (\textit{KNN}). 

Despite its overall feasibility, \textit{SDTree} introduces 
more noise 
due to its inaccuracy in reconstructing high-frequency distributions.
In contrast, our accurate local density estimation yields more stable convergence with fewer outliers.
Accurate \textit{KNN}
introduces substantial overhead (around 90\% in most scenes) since gathering thousands of nearest samples is costly. On the contrary, querying the STree only requires descending from the root to the leaf node. Consequently, the total overhead of querying and sampling distributions never exceeds 5\% in all our experiments.

\begin{figure}[tbp]
\centering
	\begin{minipage}{1.00\linewidth}
  			\includegraphics[width=1.0\linewidth]{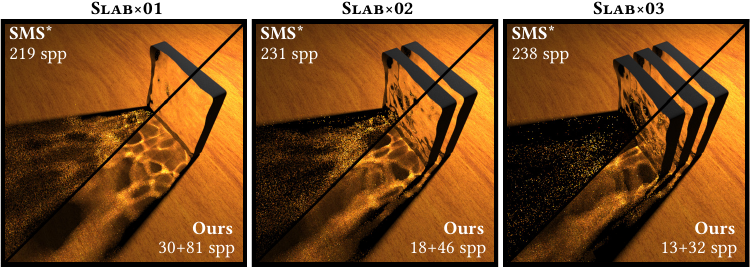} % finalized
	\end{minipage}
	\begin{minipage}{1.00\linewidth}
  			\includegraphics[width=1.0\linewidth]{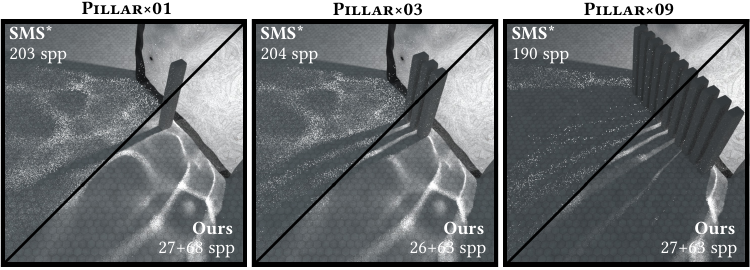} % finalized
	\end{minipage}
	\begin{minipage}{1.00\linewidth}
  			\includegraphics[width=1.0\linewidth]{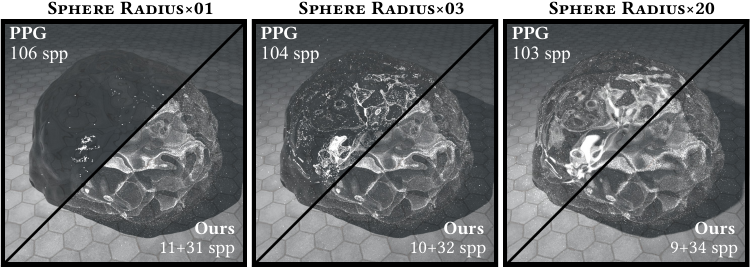} % finalized
	\end{minipage}
  \caption{\textbf{Equal-time (5 min) comparison} on scenes with varying chain lengths (top), visibility complexity (middle), and emitter radii (bottom). } 
  \Description{...}
  \label{fig_result_complex}
\end{figure}

\paragraph{Spatial filtering.}
We have employed the spatial filtering (\textit{Ours}) to improve robustness. Along with this, we include comparisons with alternative choices: no spatial filtering (\textit{NoFilter}) and \citet{mueller19guiding}'s stochastic filtering (\textit{StochasticFilter}) that randomly jitter the sample's position.
Simply disable filtering produces inferior results with visible artifacts.
Although stochastic filtering has been widely adopted in previous works \cite{Ruppert2020RobustFO, Zhu2021HierarchicalNR}, it also leads to slightly less accuracy and more fireflies.
In contrast, our approach produces results with less noise and fewer outliers with nearly no performance degradation, striking an appropriate balance between overhead and performance.

\paragraph{Spatial neighboring threshold.}

We also validate our automatic spatial neighboring threshold in Fig. \ref{fig_val_spatial_bar}. Our automatic threshold always predicts a reasonable value and generally outperforms alternative choices, including \citet{mueller19guiding}'s threshold ($4000 \cdot \sqrt{\mathrm{SPP}}$) and variants of our threshold with different coefficients.

\subsection{Impact of scene complexity}

Our method adapts well to scenes with various intricate structures. 

\paragraph{Chain length.} 
The first row of Fig. \ref{fig_result_complex} conducts a visual comparison against SMS on a series of scenes with an increasing number of transparent slabs. Since SMS neglects energy distributions during specular chain sampling, the variance increases prominently as the number of specular bounces increases. In contrast, by exploiting the continuity of specular manifolds, our method keeps a low variance even in scenes with long specular chains.

\begin{figure}[tbp]
\centering
	\begin{minipage}{1.00\linewidth}
  			\includegraphics[width=0.5\linewidth]{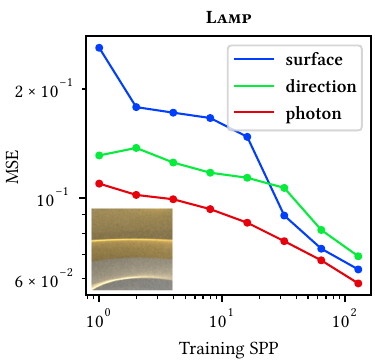} % finalized
  			\includegraphics[width=0.5\linewidth]{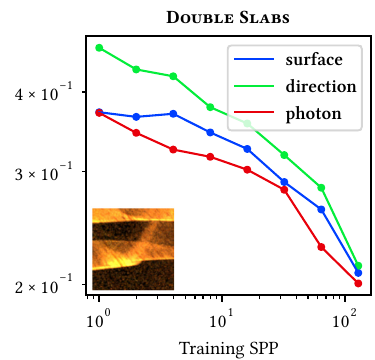} % finalized
	\end{minipage}
  \caption{\textbf{The impact of training budget and initialization strategies on the rendered image quality}. The number of training samples per pixel is given as the x-axis. Each image is finally generated with 32 spp.}
  \Description{...}
  \label{fig_plot_train}
\end{figure}

\paragraph{Visibility.} 
The second row of Fig. \ref{fig_result_complex} shows a series of scenes with increasing degrees of complex visibility stemming from multiple pillars near a transparent slab. 
SMS hardly samples solutions that pass the visibility test, while we consider visibility by reconstructing distributions from unblocked admissible chains. As a result, our method avoids high variance introduced by occlusions and works well in complex scenes.

\paragraph{Size of light sources.}
In the last row of Fig. \ref{fig_result_complex}, we present a series of scenes with various emitter sizes. 
As the size of the emitter decreases, the frequency of caustics increases. While path-guiding methods work well for large area lights, producing high-frequency caustics can be challenging. Instead, our proposed method addresses this issue by exploring admissible chains through manifold walks, allowing us to handle small area lights and complex caustic patterns effectively.

\subsection{Choice of strategies}

Our pipeline incorporates some design choices. We conduct the following experiments to validate their effectiveness.

\paragraph{Training budgets.}
Fig. \ref{fig_plot_train} shows the influence of the training budget on the rendering with the same sample rate. Generally, the variance reduces significantly as the number of training samples increases. This indicates that the training process gradually makes the sampling distribution optimal. 

\paragraph{Initialization.} 
We also compare different initialization strategies in Fig. \ref{fig_plot_train}. Here, \textit{surface} means uniformly sampling $\mathbf{x}_1$ on specular surfaces, \textit{direction} means uniformly choosing $\mbomega_D$ on the hemisphere, and \textit{photon} means reconstructing the initial distribution from 250,000 photons traced by a photon mapper. The comparison shows that initialization using photons leads to better results at the early stage of training. However, after convergence, different initialization strategies perform similarly. This shows that our method is robust to the initial state.

\begin{table*}
  \caption{Rendering statistics of our experiments. }
  \label{tab_stats}
  \begin{tabular}{llcrrrrrrrrr}
    \toprule
    Scene & Figure &  Chain types & \multicolumn{2}{c}{Budget (min.)} & \multicolumn{2}{c}{Sample per pixel} & \multicolumn{3}{c}{Time on specular (min.)} & \#Sub-path \\
    & & (dominant) & Total & Training  & Training & Rendering & Guide & PDF & Total & samples \\
    \midrule
    \textsc{Glass} & Fig. \ref{fig:teaser} & $TT, TTTT, TTTTTT$ & 240.0 & 72.0 & 80 & 195 & 3.592 & 77.886 & 215.328 & 12 519 949  \\
    \textsc{Lamp} & Fig. \ref{fig_result_2} & $T, RT$ & 5.0 & 1.5 & 46 & 116 & 0.233 & 0.388 & 3.377 & 1 347 096 \\
    \textsc{Flower} & Fig. \ref{fig_result_2} & $TT, TTR$ &  5.0 & 1.5 & 16 & 134 & 0.053 & 1.971 & 3.983 & 1 189 876 \\
    \textsc{Double Slabs} & Fig. \ref{fig_result_2} & $TT, TTTT$ & 5.0 & 1.5 & 24 & 59 & 0.096 & 1.433 & 4.127 & 2 066 782 \\ 
    \textsc{Water} & Fig. \ref{fig_result_gow} & $TTTT, TTTTTT$ & 240.0 & 72.0 & 41 & 99 & 2.621 & 223.176 & 236.883 & 1 667 185 \\ % finalized                                                                                                    
  \bottomrule
\end{tabular}
\end{table*}
\subsection{Performance analysis}

In Table \ref{tab_stats}, we report the rendering statistics of several test scenes used in this paper. As seen, for each scene, 50\%-98\% of the time is spent on specular chain sampling. In particular, a small amount of time (no more than 5\% in all the scenes) is spent on guiding (i.e., querying the spatial and directional distributions). In complex scenes, the reciprocal probability estimation has a non-negligible run-time cost since many trials are required. 

\begin{figure}[tbp]
\centering
	\begin{minipage}{1.0\linewidth}
  			\includegraphics[width=1\linewidth]{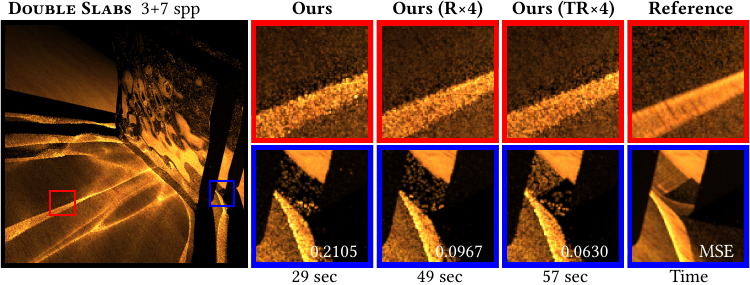}
	\end{minipage}
  \caption{\ff{The variance of reciprocal probability estimation can be reduced by repeating the process multiple times (4 in this example) and using the average, either in the rendering phase only (R×4) or both training and rendering phases (TR×4). This example also showcases the notable overhead of the estimation process.}
  } 
  \Description{...}
  \label{fig_result_repber}
\end{figure}

Our implementation needs to store the original sample, which could cause relatively heavy storage. Our consideration is twofold. On the one hand, the storage required for each raw sample is only 40 bytes. On the other hand, it is worthy of being done because the local density estimation is more accurate than existing online fitting approaches when faced with near-delta radiance distributions.

\section{Discussion and Future Work}

There are several challenges not yet solved in our method.

\paragraph{Overhead and variance of reciprocal probability estimation.} \ff{Our importance sampling works as a promising variance reduction technique, but it only considers the variance of the sampling process itself. We still rely on repeating the whole chain sampling process to find the reciprocal probability of an admissible chain. Unfortunately, the overhead and variance may be high in complex scenes where there are many solutions with small convergence basins for a configuration. Figure \ref{fig_result_repber} is an example showing the overhead and visual impact of the variance of reciprocal probability estimation. While the noise of the estimation visibly affects the image, how to balance the variance and overhead remains a challenge for future research.}

\begin{figure}[tbp]
\centering
	\begin{minipage}{1.0\linewidth}
  			\includegraphics[width=1\linewidth]{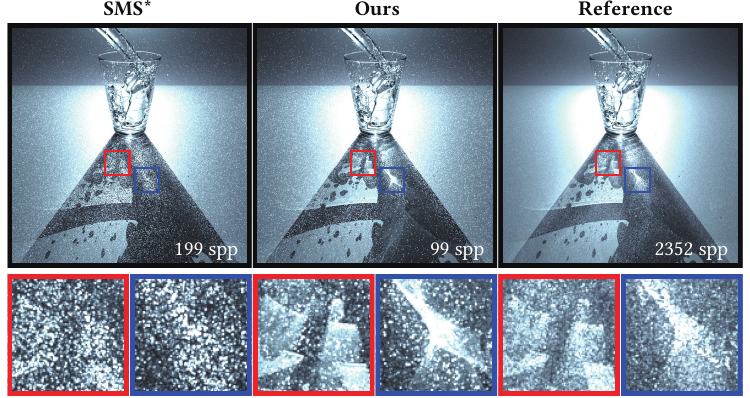}
	\end{minipage}
  \caption{\textbf{Equal-time (4 hours) rendering of the \textsc{Water} scene, demonstrating a failure case for our approach due to tiny caustic casters} (many small droplets in this scene). 
  } 
  \Description{...}
  \label{fig_result_gow}
\end{figure}

\paragraph{Online \ff{training} pitfalls} 
The training process relies on an initialization sampling strategy. This lead to a general issue that too complex or too large specular surfaces, as well as light sources, will make the training process slow. 
Insufficient training will lead to fireflies or temporal instability. 
One case is demonstrated in Fig. \ref{fig_result_gow}.

However, this is a general stubborn problem for rendering algorithms involving MC sampling. Tiny geometries make the path space complicated and hard to explore by sampling without geometry awareness \cite{Otsu2018GeometryawareML}. Using a deterministic strategy to acquire sampling distributions may alleviate this issue. 
Outlier removal \cite{Reibold2018SelectiveGS}, or biased variants \cite{Zeltner20}, can also be used in practice.

\section{Conclusion}
We have studied the problem of importance sampling specular chains that involve multiple consecutive specular bounces. Existing methods often overlook energy distributions in the discrete admissible chain space, resulting in high variance. 
To solve this problem, we have developed a comprehensive approach incorporating a specially designed continuous space for seed chain sampling. Following that, we have implemented a practical pipeline that leverages manifold path guiding to explore all the specular chains in a given scene. Additionally, our findings suggest that seed chain sampling has the potential to address the longstanding unbiased caustics rendering problem in a practical setting.

Our work is the first study focusing on importance sampling strategies for multi-bounce specular light transport.
We hope this work could promote new research interests in the Monte Carlo simulation of challenging light transport.

\begin{acks}
We would like to thank the anonymous reviewers for their valuable suggestions. 

The glass model in the scene \textsc{Glass} has been created by the cgtrader user DeepDreamDimension, and the texture in the scene \textsc{Lamp} is from the cgtrader user mehran1989. The scene \textsc{Flower} has been created by the cgtrader user Sirriarikan. The scene \textsc{Sphere}, \textsc{Slab}, \textsc{Ring}, \textsc{Double Slabs}, \textsc{Pool}, \textsc{Plane}, \textsc{RingAndSlab} and \textsc{Pillar} are modified from the test scenes of \cite{Zeltner20}. We also thank the following people for providing the models or scenes that appear in our figure: Wig42 (\textsc{Living}), UP3D (\textsc{Lamp}), Delatronic (\textsc{Stone}), and aXel (\textsc{Water}). These scenes are acquired from \cite{resources16} with modifications. Besides, our test scenes use textures from CC0 Textures and cgbookcase. We also use environment maps courtesy of HDRI Haven and Paul Debevec.

This work was supported by the National Natural Science Foundation of China (No. 61972194 and No. 62032011) and the Natural Science Foundation of Jiangsu Province (No. BK20211147).
\end{acks}


\end{document}